\documentclass{article}
\usepackage{latexsym}
\newcommand {\ra}{\rightarrow}

\newcommand {\bs}{\begin{sloppypar}}
\newcommand {\es}{\end{sloppypar}}

\newtheorem{myth}{Theorem}
\newtheorem{mylma}{Lemma}[section]

\newtheorem{mypro}{Proposition}
\newtheorem{myex}{Example}

\begin{document}
\bibliographystyle{plain}

\title{A Modal Logic Framework for  Multi-agent Belief Fusion
\thanks{A preliminary version of the paper has appeared in \cite{liau}.}}
\author{Churn-Jung Liau\\
Institute of Information Science\\ Academia Sinica,Taipei,
Taiwan\\ E-mail: liaucj@iis.sinica.edu.tw}
\date{}
\maketitle
\begin{abstract}
This paper is aimed at providing a uniform framework for reasoning
about beliefs of multiple agents and their fusion. In the first
part of the paper, we develop logics for reasoning about
cautiously merged beliefs of agents with different degrees of
reliability. The logics are obtained by combining the multi-agent
epistemic logic and multi-sources reasoning systems. Every
ordering for the reliability of the agents is represented by a
modal operator, so we can reason with the merged results under
different situations. The fusion is cautious in the sense that if
an agent's belief is in conflict with those of higher priorities,
then his belief is completely discarded from the merged result. We
consider two strategies for the cautious merging of beliefs. In
the first one, if inconsistency occurs at some level, then all
beliefs at the lower levels are discarded simultaneously, so it is
called level cutting strategy. For the second one, only the level
at which the inconsistency occurs is skipped, so it is called
level skipping strategy. The formal semantics and axiomatic
systems for these two strategies are presented. In the second
part, we extend the logics both syntactically and semantically to
cover some more sophisticated belief fusion and revision
operators. While most existing approaches treat belief fusion
operators as meta-level constructs, these operators are directly
incorporated into our object logic language. Thus it is possible
to reason not only with the merged results but also about the
fusion process in our  logics. The relationship of our extended
logics with the conditional logics
of belief revision is also discussed.
\end{abstract}
\section{Introduction}
%*********************
Recently, there has been much attention on the infoglut problem in
information retrieval research due to the rapid growth of internet
information. If a keyword is input to a commonly-used search
engine, it is not unusual to get back a list of thousands of web
pages, so the real difficulty is not how to find information, but
how to find useful information. To circumvent the problem, many
software agents have been designed to do the information search
works. The agents can search through the web and try to find and
filter out information for matching the user's need. However, not
all internet information sources are reliable. Some web sites are
out-of-date, some news provide wrong information, and someone even
intentionally spreads rumor or deceives by anonymity. Thus an
important task of  information search agents is how to merge so
much information coming from different sources according to their
degrees of reliability.

In \cite{sho1}, an agent is characterized by mental attitudes,
such as knowledge, belief, obligation, and commitment. This view
of agent, in accordance with the intentional stance proposed in
\cite{den}, has been widely accepted as a convenient way for the
analysis and description of complex systems\cite{woold}. From this
viewpoint, each information provider can be considered as an agent
and the information provided by the agent corresponds to his
belief, so our problem is also that of merging beliefs from
different agents.

The philosophical analysis of these mental attitudes has motivated
the development of many non-classical logical systems\cite{gab}.
In particular, the analysis of informational attitudes, such as
knowledge and belief, has been a traditional concern of
epistemology, a very important branch of philosophy since the
ancient times. To answer the basic questions such as ``What is
knowledge?'' ''What can we know'' and ''What are the
characteristic properties of knowledge?'', some formalism more
rigorous than natural language is needed. This results in the
development of the so-called epistemic logic\cite{hint}. This kind
of logic has attracted much attention of researchers from diverse
fields such as artificial intelligence(AI), economics,
linguistics, and theoretical computer science. Among them, the AI
researchers and computer scientist have elaborated some
technically sophisticated formalisms and applied them to the
analysis of distributed and multi-agent systems\cite{fag,mey1}.

Though the original epistemic logic in philosophy is mainly about
the single-agent case, the application to AI and computer science
put its emphasis on the interaction of agents, so multi-agent
epistemic logic is urgently needed. One representative example of
such logic is proposed by Fagin et al.\cite{fag}. In their logic,
the knowledge of each agent is represented by a normal modal
operator\cite{che}, so if no interactions  between agents occur,
this is not more than a multi-modal logic. However, the most novel
feature of their logic is the consideration of common knowledge
and distributed knowledge among a group of agents. While common
knowledge is the facts that everyone knows, everyone knows that
everyone knows, everyone knows that everyone knows that everyone
knows, and so on, distributed knowledge is that can be deduced by
pooling together the knowledge of everyone, so it is the latter
that really concerns the fusion of knowledge among agents.
However, the term ``knowledge'' is used in a broad sense in
\cite{fag} to cover the cases of belief and
information.\footnote{More precisely, the logic for belief is
called doxastic logic. However, here we will use the three terms
knowledge, belief, and information interchangeably, so epistemic
logic is assumed to cover all these notions.} Though it is
required that proper knowledge must be true, the belief of an
agent may be wrong, so there will be conflicts in general in the
beliefs to be merged. In this case, everything can be deduced from
the distributed beliefs due to the notorious omniscience property
of epistemic logic, so the merged result will be useless to
further reasoning.

Instead of directly put all beliefs of the agents together, there
are also many sophisticated techniques for knowledge base
merging\cite{bar3,bar2,bar1,cho,cho4,kon3,kon1,kon2,linj2,linj1,linj3,nebel,pra1,sub}.
Most of the approaches treats belief fusion operators as
meta-level constructs, so given a set of knowledge bases, this
kind of fusion operators will return the merged results. Some of
the works propose concrete  operators which can be used directly
in the fusion process, while the others stipulate the desirable
properties of  reasonable belief fusion  operators by postulates.
However, few of the approaches provides the capability of
reasoning about the fusion process. One of the few exceptions is
the work of multi-source reasoning\cite{cho}.

Multi-source reasoning is to model the fusion process of multiple
databases in a modal logic.  The context of the work is to merge a
set of databases according to a total ordering on the set to be
merged. Each database is a finite and satisfiable set of literals.
Two attitudes for merging are considered. According to the
suspicious attitude, if a database contains a literal inconsistent
with those in the databases of higher reliability, then the
database is completely discarded in the merged result. On the
other hand, according to the trusting attitude, if a literal in a
database is inconsistent with those in the databases of higher
reliability, only the literal is discarded, and other literals in
the database will be still considered if they are consistent with
those in the databases of higher reliability.

Since multi-source reasoning is modelled in a modal logic
framework, it is very suitable for the integration with epistemic
logic. The restriction here is that each database must be a set of
literals in multi-source reasoning, however, in the multi-agent
epistemic logic, it is expected that more complex compound
formulas will be believed by agents. Therefore, we have to extend
the multi-sources reasoning to the more general case. To achieve
the purpose, the distributed knowledge operators in multi-agent
epistemic logic may help. What we have to do is to adapt the
multi-agent epistemic logic so that the distributed knowledge
among a group of agents with reliability ordering can also be
defined. However, since the set of facts believed by an agent is
at least closed under classical logical equivalence, the trusting
attitude does not work here. For example, if $p$ and $q$ are both
believed by an agent and $\neg p\vee\neg q$ is believed by another
agent with higher reliability, then by trusting attitude, one of
$p$ or $q$ should be in the merged result (assume there do not
exist other conflicts), however, it is obvious that the belief of
the first agent is equivalent to $p\wedge q$ and if it is
expressed in this way, then no belief of the first agent (except
the obvious tautology) should be included in the merged belief.
Thus, we will only consider the merging of beliefs according to
the suspicious attitude, so this approach is very cautious from
the viewpoint of belief fusion. However, we will show that the
fusion according to the trusting attitude can also be simulated in
our logic, though the simulation is syntax-dependent. We consider
two strategies for the cautious merging of beliefs. In the first
one, if inconsistency occurs at some level, then all beliefs at
the lower levels are discarded simultaneously, so it is called
level cutting strategy. For the second one, only the level at
which the inconsistency occurs is skipped, so it is called level
skipping strategy.

The logics integrate multi-source reasoning into multi-agent
epistemic logic, so it enhance the reasoning capability of the
latter. However, since the fusion technique used in the logics is
essentially the so-called base revision in \cite{nebel}, it is too
cautious in some cases. Thus we would also like to consider the
extension of the logics with some more sophisticated fusion
operators proposed in the literatures. We show that the
multi-agent epistemic logic framework can accommodate these belief
fusion operators to  a large extent both syntactically and
semantically. This means that the belief fusion operators as a
standard add-on of  multi-agent epistemic logic should be
expectable.

The rest of the paper is organized as follows. In the next
section, the multi-agent epistemic logic  and  multi-sources
reasoning are reviewed. Then the logics integrating cautious
fusion into multi-agent epistemic logic are presented. The level
cutting and skipping strategies are presented respectively in
section \ref{sec3} and \ref{sec4}. The syntax, semantics, and
axiomatic systems of the logics will be given. In section
\ref{sec5}, the basic logics are compared with their ancestors and
another cautious inconsistency handling technique.  In section
\ref{sec6} and \ref{sec7} , accompanied by the brief introductions
of some of the most important belief fusion or revision
techniques, the possible extensions of our basic logics for
accommodating them are presented. Finally, some further research
directions are discussed in the concluding section.

\section{Logical Preliminary}
In this section, we review the syntax, semantics and some
notations for multi-agent epistemic logic and multi-sources
reasoning.
\subsection{Multi-agent epistemic logic}
In \cite{fag}, some variants of epistemic logic systems are
presented. The most basic one with distributed belief is called
K$_n^D$ by following the naming convention in \cite{che}, with $n$
being the number of agents and $D$ denoting the distributed belief
operators. In the system, no properties except logical omniscience
are imposed on the agents' beliefs. Nevertheless, in the
following, we will assume the belief of each individual agent is
consistent though the collective ones of several agents may be
not, so the system will be KD$_n^D$ where the additional axiom D
is added to K$_n^D$ for ensuring the consistency of each agent's
belief.\footnote{Though it is well accepted that KD45$_n^D$ is
more appropriate for modeling of belief with positive and negative
introspection (axioms 4 and 5), we adopt the KD$_n^D$ system for
emphasizing the agents may represent databases and their beliefs
may be just the facts stored in the databases and their
consequences.}

Assume we have $n$ agents and a set $\Phi_0$ of countably many
atomic propositions, then the set of well-formed formulas(wff) for
the logic KD$_n^D$ is the least set containing $\Phi_0$ and closed
under the following formation rules:\footnote{In \cite{fag}, the
modal operators are denoted by $K_i$ instead of $B_i$}
\begin{itemize}
\item if $\varphi$ is a wff, so are $\neg\varphi$, $B_i\varphi$,
 and $D_G\varphi$ for all $1\leq i,j\leq n$ and nonempty $G\subseteq\{1,\ldots,n\}$,
and
\item if $\varphi$ and $\psi$ are wffs, then $\varphi\vee\psi$ is,
too.
\end{itemize}
As usual, other classical Boolean connectives $\wedge$ (and),
$\supset$ (implication), $\equiv$ (equivalence), $\top$
(tautology), and $\bot$ (contradiction) can be defined as
abbreviations.

The intuitive meaning of $B_i\varphi$ is ``The agent $i$ believes
$\varphi$.'', whereas that for $D_G\varphi$ is ``The group of
agents $G$ has distributed belief $\varphi$.''. The
possible-worlds semantics provides a general framework for the
modeling of knowledge and belief\cite{fag}. In the semantics, an
agent's belief state corresponds to the extent to which he can
determine what world he is in. In a given world, the belief state
determines the set of worlds that the agent considers possible.
Then an agent is said to believe a fact $\varphi$ if $\varphi$ is
true in all worlds in this set. Since the distributed belief of a
group is the result of pooling together the individual beliefs of
its members, this can be achieved by intersecting the sets of
worlds that each agent in the groups considers possible.

Formally, a KD$_n^D$ model is a tuple $(W,({\cal B}_i)_{1\leq
i\leq n},   V)$, where
\begin{itemize}
\item $W$ is a set of possible worlds,
\item ${\cal B}_i\subseteq W\times W$ is a serial binary relation on $W$ for $1\leq i\leq n$,
\footnote{A relation $\cal R$ on $W$ is serial if $\forall w
\exists u {\cal R}(w,u)$.}
\item $  V:\Phi_0\ra 2^W$ is a truth assignment mapping each atomic
proposition to the set of worlds in which it is true.
\end{itemize}

In the following, we will use some standard notations for binary
relations. If ${\cal R}\subseteq A\times B$ is a binary relation
between $A$ and $B$, we will write ${\cal R}(a,b)$ for $(a,b)\in
{\cal R}$ and ${\cal R}(a)$ for the subset  $\{b\in B\mid {\cal
R}(a,b)\}$. Thus for any $w\in W$, ${\cal B}_i(w)$ is a subset of
$W$. Informally, ${\cal B}_i(w)$ is the set of worlds that agent
$i$ considers possible under $w$ according to his belief. The
informal intuition is reflected in the definition of satisfaction
relation. Let $M=(W, ({\cal B}_i)_{1\leq i\leq n},   V)$ be a
KD$_n^D$ model and $\Phi$ be the set of wffs, then the
satisfaction relation $\models_M\subseteq W\times\Phi$ is defined
by the following inductive rules(we will use the infix notation
for the relation and omit the subscript $M$ for convenience):
\begin{enumerate}
\item $w\models p$ iff $w\in  V(p)$ for any $p\in\Phi_0$,
\item $w\models\neg\varphi$ iff $w\not\models\varphi$,
\item $w\models \varphi\vee\psi$ iff $w\models\varphi$ or
$w\models\psi$,
\item $w\models B_i\varphi$ iff for all $u\in {\cal B}_i(w)$,
$u\models\varphi$,
\item $w\models D_G\varphi$ iff for all $u\in \bigcap_{i\in G}{\cal B}_i(w)$,
$u\models\varphi$.
\end{enumerate}

The notion of validity is defined from the satisfaction relation.
A wff $\varphi$ is valid in $M$, denoted by $\models_M\varphi$, if
for every $w\in W$, $w\models_M\varphi$, and valid in a class of
models $\cal M$, written as $\models_{\cal M}\varphi$, if for all
$M\in{\cal M}$, $\models_M\varphi$.

\subsection{Multi-sources reasoning}\label{ss1}
The context of multi-sources reasoning is the merging of $n$
databases. To encode the degrees of reliability of these
databases, the total ordering on a subset of $\{1,\ldots,n\}$ is
used. Let ${\cal TO}_n$ denote the set of all possible total
orders on the subsets of $\{1,\ldots,n\}$, $\Phi_0$ denote a
finite set of atomic propositions and ${\cal L}(\Phi_0)$ be the
classical propositional language formed from $\Phi_0$, then the
set of wffs for logic FU$_n$ (originally called FUSION in
\cite{cho}) is the least set containing $\Phi_0$ and
$\{[O]\varphi: \varphi\in{\cal L}(\Phi_0), O\in {\cal TO}_n\}$ and
being closed under Boolean connectives. If $O$ is the ordering
$i_1>i_2>\cdots>i_m$ for some
$\{i_1,\ldots,i_m\}\subseteq\{1,\ldots,n\}$, then the wff
$[O]\varphi$ means that $\varphi$ holds after merging the
databases $i_1,\ldots,i_m$ according to the specified ordering. In
this case, $O>i_{m+1}$  denotes $i_1>i_2>\cdots>i_m>i_{m+1}$.
Furthermore, the set $\{i_1,i_2,\ldots,i_m\}$ is called the domain
of $O$ and is denoted by $\delta(O)$.

Let $Lit(\Phi_0)$ denote the set of literals in ${\cal
L}(\Phi_0)$.\footnote{A literal is an atom or a negated atom.} In
the context of multi-sources reasoning, assume $DB_1,\ldots, DB_n$
are $n$ databases, where each $DB_i$ is a finite satisfiable
subset of $Lit(\Phi_0)$, then the informal semantics for the
merging databases can be given according to two attitudes. For the
suspicious attitude, only the case of $n=2$ is given in
\cite{cho}, where the definition of $DB_{1>2}$ is defined by
\[DB_{1>2}=\left\{\begin{array}{ll}
DB_1\cup DB_2 & {\rm if} \; DB_1\cup DB_2 \; {\rm is} \; {\rm
consistent},\\ DB_1 & {\rm otherwise}.
\end{array}
\right.\] On the other hand, for the trusting attitude, the
definition of $DB_O$ is given in the following recursive formula
\[DB_{O>i} = DB_O\cup\{l\in DB_i: \overline{l}\not\in DB_O\},\]
where $\overline{l}$ is the complementary of $l$. Then the
intended meaning of $[O]\varphi$ is $DB_O\models_{\rm CL}\varphi$,
where CL denotes classical propositional reasoning.

Thus an FU$_n$ model is a tuple $(W,({\cal R}_i)_{1\leq i\leq n},
  V)$, where $W$ and $  V$ are as defined in KD$_n^D$ models, and
each ${\cal R}_i$ is a serial binary relation on $W$.\footnote{In
\cite{cho},it is assumed that each ${\cal R}_i$ is an equivalence
relation. However, since nested modalities are not allowed in
FU$_n$, the difference is inessential.} The clause for
satisfaction of the formula $[O]\varphi$ is then
\begin{description}
\item $w\models[O]\varphi$ iff for all $u\in {\cal R}_O(w)$,
$u\models\varphi$,
\end{description}
where ${\cal R}_O$ is defined from ${\cal R}_i$'s according to two
attitudes. For the suspicious attitude,
\[{\cal R}_{1>2}(w)=\left\{\begin{array}{ll}
 {\cal R}_1(w) & {\rm if} \; {\cal R}_1(w)\cap {\cal R}_2(w)=\emptyset,\\
{\cal R}_1(w)\cap {\cal R}_2(w)& {\rm otherwise},
\end{array}
\right.\] for all $w\in W$. For the trusting attitude, we need
some auxiliary notations. Let $f:2^W\times 2^W\ra 2^{Lit(\Phi_0)}$
be defined as
\[f(S,T)=\{l\in Lit(\Phi_0): \forall w\in S(w\models
l)\wedge\exists w\in T(w\models l)\},\] i.e., $f(S,T)$ is the set
of literals true in all worlds of $S$ and some worlds of $T$. Then
for any $w\in W$,
\[{\cal R}_{O>i}(w) = {\cal R}_O(w)\cap\{u\in W: u\models\bigwedge f({\cal R}_i(w),{\cal
R}_O(w))\}.\] Note that if each ${\cal R}_i(w)$ denotes the set of
possible worlds in which the literals in $DB_i$ are all true, then
$f({\cal R}_i(w),{\cal R}_O(w))$ is just the set $\{l\in DB_i:
\overline{l}\not\in DB_O\}$, so ${\cal R}_{O>i}(w)$ is exactly the
set of possible worlds satisfying all literals in $DB_{O>i}$.

An axiomatic system for FU$_n$ based on trusting attitude
semantics is proposed in a recent paper\cite{cho2}. One key axiom
of that system is as follows
\[[i]l\wedge\neg[O]\neg l\supset[O>i]l,\]
where $l$ is a literal. Thus a severe restriction of FU$_n$ is the
background databases $DB_i$'s can contain only literals which may
be not the case in general practice. Though from the semantic
viewpoint, there is no essential difficulty to lift the
restriction, however the key axiom is no longer valid when the
databases contain general formulas.  On the other hand, for the
suspicious semantics, the merged database in fact contains the
distributed belief of the two databases if they are consistent.
However, since distributed belief operator is not in the language
of FU$_n$, the modal operator $[O]$ can only be characterized by
the modal operators $[i]$ for $i\in \delta(O)$. Nevertheless,
unless $\varphi$ is a literal, it seems difficult (if not
impossible) to define $D_{1,2}\varphi$ in terms of the two
individual agents' belief. Thus, a natural solution to merge
general databases in the suspicious semantics is to introduce the
distributed belief operators into the language of FU$_n$. This is
exactly what we will do in the following.

\section{Level Cutting Strategy}\label{sec3}
To unify the notations from multi-agent epistemic logic and
multi-sources reasoning, we will use the  language DBF$_n^c$ (for
distributed belief fusion and cutting strategy) defined as
follows. The wffs of DBF$_n^c$ is the least set containing
$\Phi_0$ and being closed under Boolean connectives and the
following rule:
\begin{itemize}
\item if $\varphi$ is a wff, so are $[G]\varphi$ and $[O]\varphi$
for any nonempty $G\subseteq\{1,\ldots,n\}$ and $O\in{\cal TO}_n$.
\end{itemize}
When $G$ is a singleton $\{i\}$ and $O$ is the unique total order
on $\{i\}$, we will use $[i]\varphi$ to denote both $[G]\varphi$
and $[O]\varphi$. Thus $[i]\varphi$ and $[G]\varphi$ correspond
respectively to $B_i\varphi$ and $D_G\varphi$ in KD$_n^D$, so
DBF$_n^c$ is an extension of the multi-agent epistemic logic with
distributed belief operators. On the other hand, $[O]\varphi$ and
$[i]\varphi$ are precisely those in FU$_n$, so DBF$_n^c$ is also a
generalization of multi-sources reasoning system. However, note
that nested modalities are not allowed in FU$_n$, whereas this is
not restricted in DBF$_n^c$ any more. Thus, for example, we can
include a wff $[j]\varphi$ in a database $DB_i$ which means that
$DB_i$ has the information that $\varphi$ is in $j$.

Let $Q$ be a partial order on $\{1,2,\cdots,k\}$ for some $k\leq
n$ and ${\cal O}_Q$ be the set of all total orders on
$\{1,2,\cdots,k\}$ containing $Q$, then define $[Q]\varphi$ as the
abbreviation of $\bigwedge_{O\in{\cal O}_Q}[O]\varphi$. Thus the
restriction of the modalities to total orders is not essential
since a partial order can be replaced by the set of total orders
compatible with it.

For the semantics, a DBF$_n^c$ model is just a FU$_n$ model
$(W,({\cal R}_i)_{1\leq i\leq n},   V)$. The clauses for the
satisfaction of wffs are defined exactly as in FU$_n$ model in
addition to a clause for the $[G]$ operator which is the one for
distributed knowledge in KD$_n^D$. However, the relation ${\cal
R}_O$ is now defined in an inductive way:
\[{\cal R}_{O>i}(w)=\left\{\begin{array}{ll}
 {\cal R}_O(w)& {\rm if} \; \bigcap_{j\in \delta(O>i)}{\cal R}_j(w)=\emptyset,\\
{\cal R}_O(w)\cap {\cal R}_i(w)& {\rm otherwise},
\end{array}
\right.\] for any $w\in W$.  Let $O=(i_1>i_2>\cdots>i_m)$ and
define $G_j=\{i_1,i_2\ldots,i_j\}$ for $1\leq j\leq m$ and assume
$k$ is the largest $j$ such that $\bigcap_{i\in G_k}{\cal
R}_i(w)\not=\emptyset$, then we have
\[{\cal R}_O(w)=\bigcap_{i\in G_k}{\cal
R}_i(w).\] In other words, the beliefs from the agents after the
level $k$ are completely discarded in the merged result. The
rationale behind this is if belief in level $k+1$ is not
acceptable, neither any belief in a less reliable level, so this
is a very cautious attitude to belief fusion.

The notion of validity in DBF$_n^c$ is defined just as that for
KD$_n^D$. The notation $\models_{{\rm DBF}_n^c}\varphi$ denotes
that $\varphi$ is valid in all DBF$_n^c$ model and the subscript
is usually omitted if there is no confusion. The valid wffs of
DBF$_n^c$ can be captured by the axiomatic system in
Fig~\ref{lab1}.

{\begin{figure*} \framebox[170mm]{\parbox{168mm}{
\begin{enumerate}
\item Axioms:
\begin{description}
\item P: all tautologies of the propositional calculus
\item G1: $([G]\varphi\wedge[G](\varphi\supset \psi))\supset [G]\psi$
\item G2: $\neg[i]\bot$
\item G3: $[G_1]\varphi\supset[G_2]\varphi$ if $G_1\subset G_2$
\item O1:
$\neg[\delta(O>i)]\bot\supset([O>i]\varphi\equiv[\delta(O>i)]\varphi)$
\item
O2: $[\delta(O>i)]\bot\supset([O>i]\varphi\equiv[O]\varphi)$
\end{description}
\item Rules of Inference:
\begin{description}
\item R1(Modus ponens, MP):\[\begin{array}{c}
 \varphi \;\;  \varphi\supset\psi\\ \hline  \psi\end{array}\]
\item R2(Generalization, Gen): \[\begin{array}{c}\varphi\\ \hline
[G]\varphi\end{array}\]
\end{description}
\end{enumerate}
}}
\caption{The axiomatic system for DBF$_n^c$}\label{lab1}
\end{figure*}

The axioms G1-G3 and rule R2 are those for KD$_n^D$. G1 and rule
R2 are properties of knowledge for perfect reasoners. They also
are the causes of the notorious logical omniscience problem.
However, it is appropriate to describe implicit information in
this way. G2 is the requirement that the belief of each individual
agent is consistent. G3 is a characteristic property of
distributed knowledge. The larger the subgroup, the more knowledge
it possesses. In \cite{fag}, another axiom related distributed
knowledge and individual ones is added. That is,
\[D_{\{i\}}\varphi\equiv B_i\varphi,\]
however, we do not need this because we identify $[i]\varphi$ and
$[\{i\}]\varphi$ which respectively correspond  to $B_i\varphi$
and $D_{\{i\}}\varphi$  in KD$_n^D$.  The two axioms O1 and O2
define the merged belief in terms of distributed belief in a
recursive way. O1 is the case when $ \bigcap_{j\in
\delta(O>i)}{\cal R}_i(w)\not=\emptyset$, whereas O2 is the
opposite case.

The derivability in the system is defined as follows. Let
$\Sigma\cup\{\varphi\}$ be a subset of wffs, then $\varphi$ is
derivable from $\Sigma$ in the system DBF$_n^c$ , written as
$\Sigma\vdash_{{\rm DBF}_n^c}\varphi$, if there is a finite
sequence $\varphi_1,\ldots,\varphi_m$ such that every $\varphi_i$
is an instance of an axiom schema, a wff in $\Sigma$, or
obtainable from earlier $\varphi_j$'s by application of an
inference rule. When $\Sigma=\emptyset$, we simply write
$\vdash_{{\rm DBF}_n^c}\varphi$. We will drop the subscript when
no confusion occurs. We have the soundness and completeness
results for the system DBF$_n^c$.
\begin{myth}\label{thm1}
For any wff of {\rm DBF}$_n^c$, $\models\varphi$ iff
$\vdash\varphi$.
\end{myth}
{\bf Proof}: The proof of all theorems and propositions can be
found in the appendix. $\Box$

Some basic theorems can be derived from the system.
\begin{mypro}\label{prop1}
For any  $O=(i_1>i_2>\cdots>i_m)$ and
$G_j=\{i_1,i_2\ldots,i_j\}(1\leq j\leq m)$, we have
\begin{enumerate}
\item
$\vdash(\neg[G_j]\bot\wedge[G_{j+1}]\bot)\supset([O]\varphi\equiv[G_j]\varphi)$,
where the wff $[G_{j+1}]\bot$ is deleted from the antecedent when
$j=m$.
\item $\vdash([O]\varphi\wedge[O](\varphi\supset \psi))\supset [O]\psi$,
\item $\vdash\neg[O]\bot$,
\item $\begin{array}{c}\varphi\\ \hline
[O]\varphi\end{array}$.
\end{enumerate}
\end{mypro}
Proposition \ref{prop1}.1  shows that any total order can be cut
into a head and a tail according to some consistency level, and
the merged belief according to the ordering is just the
distributed belief of the agents from the head part. Proposition
\ref{prop1}.2 and \ref{prop1}.4 show that merged belief inherits
the properties of the distributed one since the former is
equivalent to the latter for the head part of the ordering.
Furthermore, Proposition \ref{prop1}.3 shows that belief fusion
keeps consistency.

\section{Level Skipping Strategy}\label{sec4}
Though level cutting strategy is useful in practice, it is
sometimes too cautious from the viewpoint of information fusion. A
less cautious strategy is to skip only the agent causing
inconsistency and continue to consider the next level. The
strategy  corresponds to the suspicious attitude of multi-sources
reasoning and has been used in belief revision by
Nebel\cite{nebel}. This strategy is easily obtained by modifying
the inductive definition of ${\cal R}_{O>i}$ as follows.
\[{\cal R}_{O>i}(w)=\left\{\begin{array}{ll}
 {\cal R}_O(w) & {\rm if} \;{\cal R}_O(w)\cap{\cal R}_i(w)=\emptyset,\\
{\cal R}_O(w)\cap {\cal R}_i(w)& {\rm otherwise},
\end{array}
\right.\] for any $w\in W$.

According to the definition, $[O>i]\varphi$ will be equivalent to
the distributed fusion of $[O]\varphi$ and $[i]\varphi$  when the
belief of $i$ is consistent with the merged belief of $O$, so to
axiomatize reasoning under the strategy, we must view $O$ as a
virtual agent and consider the distributed belief between $O$ and
$i$. However, to get a bit more general, we will consider the
distributed belief among a group of virtual agents. Thus, we
define the wffs of the logic DBF$_n^s$(for skipping strategy) as
the least set containing $\Phi_0$ and being closed under Boolean
connectives and the following rule:
\begin{itemize}
\item if $\varphi$ is a wff, so are $[\Omega]\varphi$ for any nonempty
$\Omega\subseteq{\cal TO}_n$.
\end{itemize}
When $\Omega$ is a singleton $\{O\}$, we will write $[O]\varphi$
instead $[\{O\}]\varphi$. If $\Omega=\{O_1,\ldots, O_m\}$ is such
that $|\delta(O_i)|=1$ for all $i$'s, then $[\Omega]$ is the
distributed belief operator among ordinary agents. Therefore, the
language is more general than that of DBF$_n^c$.

For the semantics, a DBF$_n^s$ model is still a DBF$_n^c$ model,
however, the satisfaction clauses for $[O]$ and $[G]$ operators
are replaced by the following
\begin{description}
\item $w\models[\Omega]\varphi$ iff for all $u\in{\cal
R}_\Omega(w), u\models\varphi$,
\end{description}
where ${\cal R}_\Omega(w)=\bigcap_{O\in\Omega}{\cal R}_O(w)$ and
${\cal R}_O$ is defined inductively at the beginning of the
section. Given this language and semantics, the valid wffs of
DBF$_n^s$ is capture by the axiomatic system in Fig~\ref{lab2}.

{\begin{figure*} \framebox[170mm]{\parbox{168mm}{
\begin{enumerate}
\item Axioms:
\begin{description}
\item P: all tautologies of the propositional calculus
\item V1: $([\Omega]\varphi\wedge[\Omega](\varphi\supset \psi))\supset [\Omega]\psi$
\item V2: $\neg[i]\bot$
\item V3: $[\Omega_1]\varphi\supset[\Omega_2]\varphi$ if $\Omega_1\subset \Omega_2$
\item O1':
$\neg[\{O,i\}]\bot\supset([\Omega\cup\{O>i\}]\varphi\equiv[\Omega\cup\{O,i\}]\varphi)$
\item O2':
$[\{O,i\}]\bot\supset([\Omega\cup\{O>i\}]\varphi\equiv[\Omega\cup\{O\}]\varphi)$
\end{description}
\item Rules of Inference:
\begin{description}
\item R1(Modus ponens, MP):\[\begin{array}{c}
 \varphi \;\;  \varphi\supset\psi\\ \hline  \psi\end{array}\]
\item R2'(Generalization, Gen): \[\begin{array}{c}\varphi\\ \hline
[\Omega]\varphi\end{array}\]
\end{description}
\end{enumerate}
}}
\caption{The axiomatic system for DBF$_n^s$}\label{lab2}
\end{figure*}
The axioms V1-V3 and rule R2' correspond to G1-G3 and R2 for
distributed belief, but now for virtual agents instead of ordinary
agents. Nevertheless, since an ordinary agent is a special case of
the virtual one, these in fact also cover G1-G3 and R2.  O1' and
O2' are axioms for describing the level skipping strategy and
correspond exactly to the inductive definition of ${\cal
R}_{O>i}$, where $\Omega$ in these two axioms denote any subset
(empty or not) of ${\cal TO}_n$. We can still have the soundness
and completeness theorem.
\begin{myth}\label{thm2}
For any wff of {\rm DBF}$_n^s$, $\models\varphi$ iff
$\vdash\varphi$.
\end{myth}
Since operator $[O]$ is a special case of $[\Omega]$, the
properties \ref{prop1}.2 and \ref{prop1}.4 hold trivially for
DBF$_n^s$. The property \ref{prop1}.3 can be easily proved by
using V2, O1' and O2'. However, it is unclear whether a
counterpart of property \ref{prop1}.1 can be given.

\section{Related works}\label{sec5}
In this section, some important works related to the
above-mentioned logical systems will be investigated. In the
preceding sections, the strong dependence of our logics on
multi-sources reasoning and multi-agent epistemic logic has been
emphasized, so we will start from the comparison with them. Then
we also compare DBF$_n^c$ with the possibilistic logic approach to
inconsistency handling which is known to be  very cautious in
belief fusion\cite{ben1}.

\subsection{Multi-sources reasoning}
Since the original motivation of multi-sources reasoning is to
model database merging, we will also consider the relationship of
our logic to multi-sources reasoning in this context. In section
\ref{ss1}, it is assumed that $\Phi_0$ is finite and each $DB_i$
is a finite satisfiable subset of $Lit(\Phi_0)$. Let $CLS(\Phi_0)$
be the set of clauses in ${\cal L}(\Phi_0)$\footnote{A clause is a
disjunction of literals.}, then in FU$_n$, each $DB_i$ is
characterized by a wff
\begin{equation}\label{eq1}
\begin{array}{ccl}
\psi_i & = &\bigwedge\{[i]l: l\in DB_i\}\wedge\\ & &
\bigwedge\{\neg[i]c: c\in CLS(\Phi_0),
DB_i\not\vdash_{CL}c\},\end{array}
\end{equation} and the reasoning
problem is to decide whether the following holds:
\[\models\bigwedge_{i=1}^n\psi_i\supset[O]\varphi,\]
for some given $O$ and $\varphi\in {\cal L}(\Phi_0)$. The formula
$\psi_i$ asserts not only the explicit information in $DB_i$ but
also the default negative information about it. However, since in
our logic, no restrictions are put on the wffs in databases, this
kind of default wffs are potentially infinite, so we will only
assert a weaker form of wff. Let $G$ be a subset of
$\{1,2,\ldots,n\}$, then $G$ is consistent if $\bigcup_{i\in
G}DB_i$ is classically consistent, otherwise, it is inconsistent.
A subset $G$ is a maximal consistent agent group  if $G$ is
consistent and for any $i\not\in G$, $G\cup\{i\}$ is inconsistent.
Let $MCAG$ denote the class of all maximal consistent agent groups
and redefine $\psi_i=\bigwedge\{[i]\varphi: \varphi\in DB_i\}$,
then we can define the formula $\psi$ representing the databases
as
\[\psi=\bigwedge_{i=1}^n\psi_i\wedge\bigwedge_{G\in
MCAG}\neg[G]\bot.\] Thus the reasoning problem in our logic is to
decide whether $\vdash\psi\supset[O]\varphi$ holds in our system
for some given $O$ and $\varphi$. Let us use an example to
illustrate the application.
\begin{myex}
{\rm Assume there are four databases $DB_1=\{p\}, DB_2=\{q\},
DB_3=\{\neg p\vee\neg q\}$, and $DB_4=\{r,s\}$, where $p, q, r$,
and $s$ are propositional symbols, then according to the above
discussion,
\[\psi=[1]p\wedge[2]q\wedge[3](\neg p\vee\neg
q)\wedge[4](r\wedge s)
\wedge\neg[\{1,2,4\}]\bot\wedge\neg[\{1,3,4\}]\bot\wedge\neg[\{2,3,4\}]\bot.\]
By using level cutting strategy, we have the following reasoning
steps:
\[\begin{array}{ll}
1. \psi\supset(\neg[\{1,2\}]\bot\wedge[\{1,2,3\}]\bot) & G1, G3,
P, MP\\
2.(\neg[\{1,2\}]\bot\wedge[\{1,2,3\}]\bot)\supset([1>2>3>4]\varphi\equiv[\{1,2\}]\varphi)
& Prop \ref{prop1}.1\\ 3.
\psi\supset([1>2>3>4]\varphi\equiv[\{1,2\}]\varphi) & 1, 2, MP
\end{array}\]
Thus, by  epistemic reasoning in KD$_n^D$, we have the results
$\vdash\psi\supset[O](p\wedge q)$ but
$\not\vdash\psi\supset[O](r\wedge s)$ when $O=1>2>3>4$. This means
that both databases $DB_3$ and $DB_4$ are discarded according to
the ordering even only $DB_3$ is in conflict with $DB_1$ and
$DB_2$.

On the other hand, if the level skipping strategy is adopted. Then
we have the following proof.
\[\begin{array}{ll}
1. \psi\supset\neg[\{1,2\}]\bot & V3\\
2.\psi\supset([\{1>2,3\}]\bot\equiv[\{1,2,3\}]\bot) & O1', 1, P,
MP\\ 3.\psi\supset([\{1>2,4\}]\bot\equiv[\{1,2,4\}]\bot) & O1', 1,
P, MP\\ 4.\psi\supset[\{1>2,3\}]\bot & V1, V3, 2, P, MP\\
5.\psi\supset\neg[\{1>2,4\}]\bot & 3, P, MP\\6.
[\{1>2,3\}]\bot\supset([\{1>2>3,4\}]\bot\equiv[\{1>2,4\}]\bot)&
O2'\\ 7.\psi\supset\neg[\{1>2>3,4\}]\bot& 4, 5, 6, P, MP\\
8.\psi\supset([\{1>2,4\}]\varphi\equiv[\{1,2,4\}]\varphi)& 1, O1',
P, MP\\
9.\psi\supset([\{1>2>3,4\}]\varphi\equiv[\{1>2,4\}]\varphi)& 4,
O2', P, MP\\
10.\psi\supset([1>2>3>4]\varphi\equiv[\{1>2>3,4\}]\varphi)& 7,
O1', P, MP\\
11.\psi\supset([1>2>3>4]\varphi\equiv[\{1,2,4\}]\varphi) & 8, 9,
10, P, MP
\end{array}\]
Thus we have $\vdash\psi\supset[O](p\wedge q\wedge r\wedge s)$ by
epistemic logic, i.e. only $DB_3$ is discarded for its conflict
with $DB_1$ and $DB_2$. $\Box$}
\end{myex}

The reasoning in the above example corresponds to the suspicious
attitude in merging databases. In \cite{liau}, it is shown that
the trusting attitude merging can also be simulated in the system
DBF$_n^s$, though the simulation is somewhat awkward. While the
simulation in \cite{liau} is  restricted to the databases
containing only literals, here we consider the general case.

To simulate the trusting attitude merging, recall that for a
partial order $Q$ and the set ${\cal O}_Q$ of all total orders
compatible with it, $[Q]\varphi$ is the abbreviation of
$\bigwedge_{O\in{\cal O}_Q}[O]\varphi$. The basic idea of the
simulation is to split each database containing $m$ wffs into $m$
sub-databases, so we have in total $\sum_{i=1}^{n}|DB_i|$
sub-databases. Let $DB_{ij}$ denote the $j$-th sub-database
obtained from the $i$-th database, then a total ordering
$O\in{\cal TO}_n$ is transformed into a partial ordering $Q$ on
the set ${\cal ID}=\{ij\mid DB_{ij}$ is a sub-database$\}$ such
that $i_1j_1>i_2j_2$ in $Q$ iff $i_1>i_2$ in $O$. Then the
databases are represented by the following wff
\[\psi'=\bigwedge_{ij\in{\cal ID}}\psi_{ij}\wedge\bigwedge_{G\in
MCAG'}\neg[G]\bot\] where  $\psi_{ij}=\bigwedge\{[ij]\varphi\mid
\varphi\in DB_{ij}\}$ and $MCAG'$ is the class of all maximal
consistent agent subgroups of ${\cal ID}$. Thus, to decide whether
$\varphi$ is derivable from the merging of $DB_1,DB_2,\cdots,
DB_n$ according to a total order $O$ by the trusting attitude, we
only have to do the following deduction in DBF$_n^s$.
\[\vdash\psi'\supset[Q]\varphi\]

The idea is illustrated in the following example.
\begin{myex}
{\rm Assume there are two databases $DB_1=\{p\vee q\}$ and
$DB_2=\{\neg p, \neg q\}$, then according to the reasoning in
DBF$_n^c$ or DBF$_n^s$, we have
$\vdash\psi\supset([1>2]\varphi\equiv[1]\varphi)$, where
$\psi=[1](p\vee q)\wedge[2]\neg p\wedge[2]\neg q$. Thus $DB_2$ is
completely discarded in the merging process. However, if we first
split $DB_2$ into two sub-databases $DB_{21}=\{\neg p\}$ and
$DB_{22}=\{\neg q\}$ and let $DB_{11}=DB_1$, then we have
$\vdash\psi'\supset([O_1](\neg p\vee\neg q)\wedge[O_2](\neg
p\vee\neg q)$ where $O_1= 11>21>22$, $O_2=11>22>21$, and
$\psi'=[11](p\vee q)\wedge[21]\neg p\wedge[22]\neg
q\wedge\neg[\{11,21\}]\bot\wedge\neg[\{11,22\}]\bot\wedge\neg[\{21,22\}]\bot$.
In other words, $\neg p\vee\neg q$ will be derivable from the
merging results according to the ordering $1>2$ in the original
databases. Note that here $\psi$ and $\psi'$ are different wffs
since they use different modal operators, though they essentially
represent the same database contents. It must also be noted that
the simulation is syntax-dependent since if the original second
database is given as $\{\neg p\wedge\neg q\}$ which is equivalent
to $DB_2$, then we can not split it any more. }$\Box$
\end{myex}

\subsection{Multi-agent epistemic reasoning}
Obviously, both DBF$_n^c$ and DBF$_n^s$ are conservative
extensions of KD$_n^D$ in the sense that if we uniformly replace
the modal operators $B_i$ and $D_G$ in a wff $\varphi$ of KD$_n^D$
by $[i]$ and $[G]$ respectively, then $\models_{KD_{n}^D}\varphi$
iff the replaced wff is valid in DBF$_n^c$ or DBF$_n^s$. Thus our
systems can do all reasoning that KD$_n^D$ can. Furthermore, if
some additional axioms are added, we can turn our systems into
conservative extensions of other epistemic logic systems. For
example, if the following two axioms 4 and 5 are added, then our
systems can do the reasoning of KD45$_n^D$ system which is in
general accepted as the logic for modelling agent's beliefs.
\[4. [i]\varphi\supset[i][i]\varphi\]
\[5. \neg[i]\varphi\supset[i]\neg[i]\varphi\]
However, if an additional axiom T (called knowledge axiom):
$[i]\varphi\supset\varphi$ is added to the above-extended system,
then it will degenerate into the ordinary S5$_n^D$ system in the
sense that each wff $[O]\varphi$ is provably equivalent to
$[\delta(O)]\varphi$ since no conflicts may exist if what every
agent knows is true. In the following, let us look at some
examples of integrated reasoning about the multi-agent beliefs and
their fusion.
\begin{myex}{\rm
If a set of premises $\{\neg[\{1,2\}]\bot\vee\neg[\{1,3\}]\bot,
[1](p\supset q), [2]p, [3]\neg q\}$ is given for three agents,
then it can be derived that \[\vdash_{{\rm
DBF}_n^s}[1>2>3]((p\wedge q)\vee(\neg p\wedge\neg q))\] and
\[\vdash_{{\rm
DBF}_n^c}[1>2>3](p\supset q).\] The wff
$\neg[\{1,2\}]\bot\vee\neg[\{1,3\}]\bot$ says that if the beliefs
of agents 1 and 2 are incompatible, then those of 1 and 3 are
compatible, so the level skipping strategy will either accept the
belief of agent 2 or skip it and consequently accept that of agent
3. This example shows that we can reason with the compatibility of
the agents' beliefs in the uniform framework of epistemic
reasoning and information fusion.}$\Box$
\end{myex}

The next example shows that the belief about belief may play a
role in the fusion process.
\begin{myex}{\rm Assume there are two agents whose beliefs are
described by the following set:
\begin{eqnarray*}
& &\{[1]\neg[\{1,2\}]\bot, [1]p, [1][1]p, [1][2]q\\
&  &[2][\{1,2\}]\bot, [2]q, [2][2]q, [2][1]p\}
\end{eqnarray*}
Then it can be shown that $[1>2]p\wedge[2>1]q$, $[1][1>2](p\wedge
q)\wedge[1][2>1](p\wedge q)$, and $[2][1>2]p\wedge[2][2>1]q$ are
derivable in both DBF$_n^s$ and DBF$_n^c$. Thus the belief of
agent 1 is incorrect because he wrongly believes that he is
consistent with agent 2, while agent 2 in fact disagrees with him
on the consistency between them.}$\Box$
\end{myex}

Sometimes, it is possible to infer the beliefs of individual
agents from their merged beliefs. The next example shows a very
simple case.
\begin{myex}
{\rm Assume it is known that two premises $[1>2]p$ and $[2>1]\neg
p$ hold, then we have the following derivation in DBF$_n^c$ (where
Pre in the derivation means a premise).
\[\begin{array}{ll}
1. \neg[\{1,2\}]\bot\supset([1>2]p\supset[\{1,2\}]p) & O1\\
2. \neg[\{1,2\}]\bot\supset([2>1]\neg p\supset[\{1,2\}]\neg p) & O1\\
3. \neg[\{1,2\}]\bot\supset([\{1,2\}]p\wedge[\{1,2\}]\neg p)& {\rm
Pre}, 1, 2, P, MP\\
4. \neg[\{1,2\}]\bot\supset[\{1,2\}]\bot & 3, P, G1, MP, Gen\\
5. [\{1,2\}]\bot & 4, P\\
6. [1>2]p\supset[1]p & 5, O2, MP\\
7. [2>1]\neg p\supset[2]\neg p & 5, O2, MP\\
8. [1]p & {\rm Pre}, 6, MP\\
9. [2]\neg p & {\rm Pre}, 7, MP\\
10. [1]p\wedge[2]\neg p & 8, 9, P, MP
\end{array}\]
When there are more than two agents, the situation would become
more complicated. However, it is still possible to derive some
individual or partially merged beliefs from the totally merged
ones.}$\Box$
\end{myex}

A research area related to both  epistemic logic and belief fusion
is the modal logics for representing inconsistent beliefs. In
\cite{mey2}, an epistemic default logic is proposed for the
representation of inconsistent beliefs caused by default
reasoning. The logic is based on S5P developed in
\cite{mey3,mey4,mey5} for modelling the monotonic part of default
reasoning that deals with plausible assumptions. The basic
modalities of S5P consist of an S5 epistemic operator $K$ and a
number of K45 belief operators $P_i(1\leq i\leq n)$. A wff
$P_i\varphi$ means that $\varphi$ is a plausible working belief
according to some context or default rules. Since conflict between
default rules is not unusual, it is possible that
$P_i\varphi\wedge P_j\neg\varphi$ holds. Though $P_i$ corresponds
to an application context of some default rules, it can also be
seen as the belief operator of some agent, so in this regard, the
logic is like a multi-agent epistemic logic with an S5-based
epistemic operator for the authority. However, instead of
reasoning about the merging of different working beliefs in the
logic directly, a downward reflection approach is adopted in
\cite{mey2}. Since the $P_i$ operators are only applied to
objective wffs in \cite{mey2}, the downward reflection function
maps a set of S5P wffs (especially wffs of the form $P_i\varphi$)
into  a set of non-modal formulas. Some downward reflection
mechanisms are employed to resolve the inconsistency between
working beliefs of different contexts. The one based on the
explicit ordering on frames is essentially similar to our cautious
merging. The main difference is that we take the orderings as
modal operators and reason about the fusion results directly in
the object language, while the downward reflection approach
consider the fusion in a meta-level.

\subsection{Inconsistency handling in possibilistic logic}
In \cite{ben1}, it is shown that the possibilistic logic approach
to database fusion is very cautious, so a natural question is how
the level cutting strategy is related with it.  Here, we shown
that the inconsistency handling technique of possibilistic logic
can be modelled in the strategy.

Possibilistic logic(PL) is proposed by Dubois and Prade for
uncertainty reasoning\cite{dp1,dp2,dp3}. The semantic basis of PL
is the possibility theory developed by Zadeh from fuzzy set
theory\cite{zad}. Given a universe $W$, a {\em possibility
distribution \/} on $W$ is a function $\pi : W\rightarrow[0,1]$.
Obviously, $\pi$ is a characteristic function of a fuzzy subset of
$W$. Two measures on $W$ can be derived from $\pi$. They are
called possibility and necessity measures and denoted by $\Pi$ and
$N$ respectively. Formally, $\Pi, N : 2^W\rightarrow[0,1]$ are
defined as
\[\Pi(A) = \sup_{w\in A}\pi(w),\]
\[N(A) = 1 - \Pi(\overline{A}),\]
where $\overline{A}$ is the complement of $A$ with respect to $W$.

In \cite{dp3}, a fragment for necessity-valued formula in PL,
called PL1, is introduced. Each wff of PL1 is of the form
$(\varphi,\alpha)$, where $\varphi\in{\cal L}(\Phi_0)$ and
$\alpha\in(0,1]$ is a real number. The number $\alpha$ is called
the {\em valuation \/} or {\em weight \/} of the formula.
$(\varphi,\alpha)$ expresses that $\varphi$ is certain at least to
degree $\alpha$. Formally, a model for PL1 is given by a
possibility distribution $\pi$ on the set $W$ of classical truth
assignments for ${\cal L}(\Phi_0)$. For any $\varphi\in{\cal
L}(\Phi_0)$, we can define $|\varphi|$ as the set of truth
assignments satisfying $\varphi$ . Then, by identifying $\varphi$
and its truth set $|\varphi|$, a PL1 model $\pi$ satisfies
$(\varphi,\alpha)$, denoted by $\pi\models(\varphi,\alpha)$, if
$N(\varphi)\geq\alpha$. Let $\Sigma=\{(\varphi_i,\alpha_i): 1\leq
i\leq m\}$ be a finite set of PL1 wffs, then $\Sigma\models_{\rm
PL1}(\varphi,\alpha)$ if for each $\pi$,
$\pi\models(\varphi_i,\alpha_i)$ for all $1\leq i\leq m$ implies
$\pi\models(\varphi,\alpha)$.  It is shown that the consequence
relation in PL1 can be determined completely by the least specific
model satisfying $\Sigma$. That is, if $\pi_\Sigma: W\ra[0,1] $ is
defined by
\[\pi_\Sigma(w)=
\min\{1-\alpha_i\mid w\models\neg\varphi, 1\leq i\leq m\},\] where
$\min\emptyset=1$, then $\Sigma\models_{\rm PL1}(\varphi,\alpha)$
iff $\pi_\Sigma\models(\varphi,\alpha)$.

A special feature of PL1 is its capability to cope with partial
inconsistency. For $\Sigma$ defined as above, let $\Sigma^*$
denote the set of classical formulas $\{\varphi\mid 1\leq i\leq
m\}$. Then the set $\Sigma$ is said to be partially inconsistent
when $\Sigma^*$ is classically inconsistent. It can be easily
shown that $\Sigma$ is partially inconsistent iff $\sup_{w\in
W}\pi_\Sigma(w)<1$.  Thus $\sup_{w\in W}\pi_\Sigma(w)$ is called
the consistency degree of $\Sigma$, denoted by $Cons(\Sigma)$, and
$1-Cons(\Sigma)$ is called the inconsistency degree of $\Sigma$,
denoted by $Incons(\Sigma)$. When $\Sigma$ is partially
inconsistent, it can be shown that $\Sigma\models_{\rm PL1}(\bot,
Incons(\Sigma))$, so for any classical wff $\varphi$, $(\varphi,
Incons(\Sigma))$ is a trivial logical consequence of $\Sigma$. On
the contrary, if $\Sigma\models_{\rm PL1}(\varphi,\alpha)$ for
some $\alpha>Incons(\Sigma)$, then $\varphi$ is called a
nontrivial consequence of $\Sigma$.

To model the nontrivial deduction of PL1, we assume that the
weights of the wffs are drawn from a finite subset ${\cal
V}=\{\alpha_1,\ldots,\alpha_n\}$ of $(0,1]$. Without loss of
generality, we can assume $\alpha_1>\cdots>\alpha_n$. Let us
define $n$ databases from $\Sigma$ as
$DB_i=\{\varphi\mid(\varphi,\alpha_i)\in\Sigma\}$ for $1\leq i\leq
n$. It can easily be seen that when each $DB_i$ is classically
consistent, then for any $\varphi\in{\cal L}(\Phi_0)$, $\varphi$
is a nontrivial consequence of $\Sigma$ iff $\vdash_{{\rm
DBF}_n^c}\psi\supset[1>2>\cdots>n]\varphi$, where $\psi$ is the
formula representing the databases.

\section{Incorporating Other Fusion Operators}\label{sec6}
While we adopt a modal logic approach to belief fusion, there have
been also a lot of works on knowledge merging by using  meta-level
operators\cite{bar3,bar2,bar1,kon3,kon1,kon2,linj2,linj1,linj3,pra1,sub}.
In the meta-level approach, a merging operator is in general used
to combine a set of knowledge bases $T_1, T_2,\cdots, T_k$, where
each knowledge base is a theory in some logical langauge. The main
difference between our approach and theirs is that the belief
fusion operators are incorporated into the object language in our
logic, so we can reason not only with the merged results but also
about the fusion process. However, the suspicious attitude used in
our logic may be too cautious in some cases. Thus  our logic
should also be extended to accommodate these more sophisticated
knowledge merging operators both syntactically and semantically.
In the following, we will describe these operators briefly and
discuss some possible extensions of our logic for incorporating
them into the modal language.

In the presentation below, we will extensively use the notions of
pre-order. Let $S$ be a set, then a pre-order over $S$ is a
reflexive and transitive binary relation $\leq$ on $S$. A
pre-order over $S$ is called total (or connected) if for all
$x,y\in S$, either $x\leq y$ or $y\leq x$ holds. We will write
$x<y$ as the abbreviation of $x\leq y$ and $y\not\leq x$. For a
subset $S'$ of $S$, $\min(S',\leq)$ is defined as the set $\{x\in
S'\mid\forall y\in S', y\not<x\}$.

\subsection{Combination by maximal consistency} One of the
earliest approaches to knowledge  merging is to manipulate the
maximal consistent subsets of the union of the component
databases. In \cite{bar3,bar2,bar1},  knowledge bases with
integrity constraints are combined by a meta-level combination
operator to form a new knowledge base. While in \cite{bar2,bar1},
logic programs and default logic theories are considered which
have different semantics than the classical logic, the basic idea
for combining first-order theories in \cite{bar3} can be carried
out in our logic. In \cite{bar3}, a combination operator $C$ maps
a set of knowledge bases $\{T_1,\cdots, T_k\}$ and a set of
integrity  constraints $IC$ into a new knowledge base
$C(T_1,\cdots,T_k, IC)$ which can be roughly considered as the
disjunction of maximally consistent subsets of $T_1\cup
T_2\cup\cdots\cup T_k$ with respect to $IC$.

Unlike our fusion operators which correspond to total orders on
the agents, the combination operator assumes all knowledge bases
are equally important, so there are no priorities among them,
though the priority is obviously given to the integrity
constraints. Therefore, by using the partial order fusion
operators, we can analogously model the combination operator in
our logic. Let us consider $n$ agents where the belief of agent 1
is the set $IC$ and each sentence in $T_1\cup T_2\cup\cdots\cup
T_k$ is exactly represented as the belief of  one agent in
$\{2,\cdots, n\}$, then for the partial order
$Q=\{1>2,1>3,\cdots,1>n\}$, the modal operator $[Q]$ can produce
the same result as the combination operator $C$. Note that just
like the simulation of trusting attitude multi-sources reasoning
in our logic, the maximally consistent combination is also
syntax-dependent.

In \cite{kon3}, it is argued that the maximally consistent
combination lacks many desirable properties of knowledge merging.
This is due to the fact that the source of information is lost in
the combination process. Some improvements based on the selection
of some maximally consistent subsets instead of all ones are then
proposed to circumvent the problem. Three approaches are suggested
according to the difference of the selection functions. The first
selects from the set of maximally consistent subsets those
consistent with the most knowledge bases, the second selects those
that have least difference (in terms of number of sentences) with
the knowledge bases, and the third selects those that fit the
knowledge bases on a maximum number of sentences. Though these
improvements indeed satisfy the desirable logical properties
argued by the author, they are all syntactical operator and lack a
model-theoretic semantic characterization. Furthermore, since the
second and the third improvements are based on the comparison of
cardinalities of sets of wffs, they works only for finite
knowledge bases. This makes it difficult to incorporate these
improved combination operators into our logic where each agent's
beliefs are closed under logical consequence. Fortunately, there
are other elegant merging operators with the desirable logical
properties which can be incorporated into our framework, so we
will consider some of them in the following sections.

Yet another syntax-based approach is to remove the wffs causing
inconsistency.  In \cite{ben1}, this approach is explored when
only local ordering between the wffs is given. However, the
approach is more algorithmic and it seems not appropriate to
incorporate it into our framework.

\subsection{Combination by meta-information}
In the combination by maximal consistency, it is assumed that no
information about how to combine the knowledge bases is available.
However, sometimes the users can provide valuable meta-information
about the combination process, such as the reliability of the
component databases, the user's preference, or the interaction of
different databases, etc. In \cite{pra1}, a kind of priorities
between sets of propositional atoms is represented and the
combination is made according to the prioritized information. In
fact, our fusion operators (either total orders or partial ones)
also encode a kind of priorities. The main difference is that our
priorities are between agents while theirs are between the sets of
propositions believed by the agents. However, since transitivity
is not required for the priority relation in \cite{pra1}, there
may exist cyclic priorities (i.e., $x>y$ and $y>x$ holds
simultaneously). In such cases, there would not be combined
knowledge bases satisfying the priorities. Furthermore, since the
knowledge bases in \cite{pra1} are just sets of propositional
atoms, the approach  applies only to  deduction-free relational
databases and lacks the capability of reasoning about the
inter-relationship between the knowledge bases.

A more flexible way for specifying the meta-information is
proposed in \cite{sub}. In that work, a set of local databases
$DB_1, \cdots, DB_n$ is combined with a supervisory knowledge base
$S$. Intuitively, $S$ contains conflict resolution information.
Since the databases are expressed in a very rich language, the
supervisory knowledge base can specify complex relations between
local databases. The language is called annotated logic and is
constructed from some base language and a set of annotations.
These annotations can denote the truth values for many-valued
logic, timestamps, uncertainties, etc., so the expressive power of
annotated logic is quite rich. In the framework, the local
databases are just sets of sentences in the annotated logic,
whereas the supervisory knowledge base contains sentences in
another annotated logic where each atom is indexed by a subset of
$\{1,2,\cdots, n, {\bf s}\}$.

To compare the framework in \cite{sub} with our logic, let us
assume the only annotations are  the classical truth values
$\{{\bf t},{\bf f}\}$, so the annotated logic reduces to the
classical one. In this simplified case, the annotations can be
simply removed and each local database contains logic program
clauses of the form
\[p_0 \leftarrow p_1 ,\cdots, p_m, {\bf
not} p_{m+1},\cdots, {\bf not}p_{m+k}\] where for all $0\leq i\leq
m+k$, $p_i$ is an atomic formula in classical logic, whereas the
supervisory knowledge base $S$ contains indexed clauses of the
form
\begin{equation}\label{eq2}p_0:\{{\bf s}\}\leftarrow
p_1:D_1,\cdots, p_m:D_m, {\bf not}(p_{m+1}:D_{m+1}),\cdots, {\bf
not}(p_{m+k}:D_{m+k})\end{equation} where for all $1\leq i\leq
m+k$, $D_i\subseteq\{1,2,\cdots, n, {\bf s}\}$. The intended
meaning of $p:D_i$ is that the databases in $D_i$ jointly say that
$p$ is true. The meaning is specified by a combination axiom
scheme which is equivalent to
\begin{equation}\label{eq3}p:D\leftarrow\bigvee_{\emptyset\subset D'\subset D}p:D'\end{equation}
in our simplified case. For each local data base $DB_i$, the
amalgamation transform of $DB_i$, $AT(DB_i)$, is defined as the
result of replacing each clause $p_0 \leftarrow p_1 ,\cdots, p_m,
{\bf not} p_{m+1},\cdots, {\bf not}p_{m+k}$ in $DB_i$ by
\begin{equation}\label{eq4}p_0:\{i\} \leftarrow p_1:\{i\} ,\cdots, p_m:\{i\}, {\bf
not} (p_{m+1}:\{i\}),\cdots, {\bf
not}(p_{m+k}:\{i\})\end{equation} Consequently, the {\em amalgam
\/} of $(DB_1,\ldots, DB_n, S)$ is defined as the amalgamated
knowledge base \[S\cup\bigcup_{i=1}^n AT(DB_i)\cup\;\mbox{\rm
Combination axioms}\].

Though the semantics of the amalgamated knowledge base is given
according to that of logic program, so not comparable with that of
classical logic. However, the idea of supervisory knowledge base
can be easily realized in the multi-agent epistemic logic (and so
in our logic). In fact, the clause in (\ref{eq2}) can be
translated into our logic as
\[\bigwedge_{1\leq i\leq m}[D_i]p_i\wedge\bigwedge_{m+1\leq i\leq
m+k}[D_i]\neg p_i\supset[{\bf s}]p_0,\] whereas the combination
axiom (\ref{eq3}) is a special case of the axioms G3 or V3 in our
systems. Though the annotated logic provides a far richer
expressive power in the representation of objective knowledge than
our systems and the supervisory knowledge base can express
conflict resolution information among local databases, the
framework in \cite{sub} still lacks the capability of reasoning
about mutual information. Contrarily, each agent can easily reason
about the beliefs of other agents in our logic. For example, it is
possible to say that agent $i$ believes that if agent $j$ believes
$\varphi$, then agent $k$ would also do. This somewhat reflects
the essential difference between the modal logic approach and the
meta-level ones to the belief fusion.

\subsection{Merging by majority}\label{sec:major}
Though the maximal consistent combination resolves the conflicts
between knowledge bases, it does not reflect the view of the
majority. For example, if three knowledge bases $T_1=\{\varphi\},
T_2=\{\varphi\}$, and $T_3=\{\neg\varphi\}$ are combined by the
maximal consistent combination rule, the result would be just a
knowledge base containing the tautology. However, if the majority
view is taken into account, then the result would be
$\{\varphi\}$. In \cite{linj2}, a merging operator  reflecting the
views of majority is proposed for knowledge bases consisting of
finite propositional sentences. Since the propositional language
is assumed finite there, the   so-called Dalal distance between
two interpretations of the language is used\cite{dal}. It is
defined as the number of atoms whose valuations differs in the two
interpretations. Let $dist(w,w')$ denote the Dalal distance
between two interpretations $w$ and $w'$, then the distance from
$w$ to a theory $T$, denoted by $dist(w,T)$, is defined as
\begin{equation}\label{eq5}
dist(w,T)=\min\{dist(w,w')\mid w'\models T\}. \end{equation} Given
a set of knowledge bases $T_1, T_2,\cdots, T_k$ to be merged, a
total pre-order $\preceq_{\{T_1, T_2,\cdots, T_k\}}$ is defined on
the set of interpretations by
\begin{equation}\label{eq6}w\preceq_{\{T_1, T_2,\cdots, T_k\}} w' \;\; {\rm iff}\;\;
\sum_{i=1}^k dist(w,T_i)\leq \sum_{i=1}^k
dist(w',T_i)\end{equation} Then the merged result $Merge(T_1,
T_2,\cdots, T_k)$ is the theory whose models are all
interpretations minimal with respect to the order $\preceq_{\{T_1,
T_2,\cdots, T_k\}}$.

In \cite{linj3}, a set of postulates for characterizing the
merging function is presented and its corresponding
model-theoretic characterization is also given. It is then shown
that  $Merge$ is indeed a function satisfying the postulates. In
\cite{linj1}, the function $Merge$ is further generalized for the
application in weighted knowledge bases. Let $wt:\{T_1,
T_2,\cdots, T_k\}\ra R^+$ is a weight function which assigns to
each component knowledge base a positive real number, then the
total pre-order in (\ref{eq6}) is changed into
\begin{equation}\label{eq7}w\preceq_{(\{T_1, T_2,\cdots, T_k\},wt)} w' \;\; {\rm iff}\;\;
\sum_{i=1}^k dist(w,T_i)\cdot wt(T_i)\leq \sum_{i=1}^k
dist(w',T_i)\cdot wt(T_i)\end{equation} Then the merged result
$Merge(T_1, T_2,\cdots, T_k, wt)$ is the theory whose models are
all interpretations minimal with respect to the order
$\preceq_{(\{T_1, T_2,\cdots, T_k\},wt)}$.

Since the weighted version of the merging function is more
general, we will consider the extension of our logic for the
weighted merging operator. First, a new class of modal operators
$[M(G, wt)]$ for any nonempty $G\subseteq\{1,2,\cdots,n\}$ and
weight function $wt: G\ra R^+$ is added to our logic language.
Then the semantics for the new modal operators is defined by
extending a possible world model to $(W, ({\cal R}_i)_{1\leq i\leq
n}, V, \mu)$, where $(W, ({\cal R}_i)_{1\leq i\leq n}, V)$ is a
DBF$_n^s$ (or DBF$_n^c$) model, whereas $\mu: W\times W\ra
R^+\cup\{0\}$ is a distance metric function between possible
worlds satisfying $\mu(w,w)=0$ and $\mu(w,w')=\mu(w',w)$.

It must be noted that our possible worlds are more than the truth
assignments of the propositional symbols, so it is inappropriate
to define the distance between two possible worlds by merely
enumerating the number of atoms whose valuations differs in the
two worlds. However, it is assumed a distance metric between
possible worlds can be defined just as in the semantics of
conditional logic\cite{nute,sm}. To give the semantics of the new
operators, we first define the distance from a possible world $w$
to the belief state of an agent $i$ in the possible world $u$ by
\begin{equation}\label{eq8}
dist_u(w,i)=\inf\{\mu(w,w')\mid (u,w')\in{\cal R}_i\}.
\end{equation}
Then a total pre-order $\preceq_{(G,wt)}^u$ on the possible worlds
is defined for each possible world $u$ and modal operator $[M(G,
wt)]$
\begin{equation}\label{eq9}w\preceq_{(G,wt)}^u w' \;\; {\rm iff}\;\;
\sum_{i\in G} dist_u(w,i)\cdot wt(i)\leq \sum_{i\in G}
dist_u(w',i)\cdot wt(i).\end{equation} The most straightforward
definition for the satisfaction of  the wff $[M(G,wt)]\varphi$ is
\[u\models[M(G,wt)]\varphi \;\; \mbox{\rm iff for all}
\;\;w\in {\cal R}_{M(G,wt)}(u), w\models\varphi,\] where ${\cal
R}_{M(G,wt)}$ is a binary relation  over the possible worlds such
that ${\cal R}_{M(G,wt)}(u)=\min(W,\preceq_{(G,wt)}^u)$. However,
since for infinite $W$, the set $\min(W,\preceq_{(G,wt)}^u)$ may
be empty, the definition may result in $u\models[M(G,wt)]\bot$ in
some cases. Alternatively, since $\preceq_{(G,wt)}^u$ is a total
pre-order, it is just like the a system-of-spheres in the
semantics of conditional logic\cite{nute}, so we can define the
satisfaction of the wff $[M(G,wt)]\varphi$ by
\[u\models[M(G,wt)]\varphi \;\; \mbox{\rm iff there exists}\;\;
w_0 \;\; \mbox{\rm such that for all}\;\;w\preceq_{(G,wt)}^u w_0 ,
w\models\varphi.\]

An alternative approach to do majority merging is to employ the
graded modal logic in \cite{mey6,hoek}. In the logic, a set of
modal operators $K_m$, where $m$ is a natural number, is in place
of the ordinary epistemic or doxastic operators. In the single
agent case, a modal formula $K_m\varphi$ means that in all
possible worlds the agent considers possible, there are at most
$m$ worlds at which $\varphi$ is false. By the abbreviations,
$M_m\varphi\equiv\neg K_m\neg\varphi$, $M!_0\varphi\equiv
K_0\neg\varphi$, and $M!_m\varphi\equiv(M_{m-1}\varphi\wedge\neg
M_m\varphi)$, it can be seen that $M!_m\top\wedge
K_{\lfloor\frac{m}{2}\rfloor}\varphi$ means the wff $\varphi$ is
true at more than half of the worlds. By generalizing this kind of
graded modal operators for distributed belief fusion, we can
consider the modal operators $[G]_r$ for any real number $r$ and
subset of agents $G$. The semantics for $[G]_r\varphi$ is
interpreted in the multi-agent epistemic logic model such that
\begin{center}
$w\models[G]_r\varphi$ iff $\frac{|\{u\models\varphi\mid
u\in\cup_{i\in G}{\cal R}_i(w)\}|}{|\cup_{i\in G}{\cal
R}_i(w)|}>r$.
\end{center}
Then $[G]_c$ for some threshold $c\geq 0.5$ can be taken as the
fusion operator which merges the beliefs of agents in $G$.
However, it must be noted that the majority considered in the
graded modal logic is the majority of possible worlds instead of
that of agents. Furthermore, by using the cardinality of sets of
possible worlds in the definition, the semantic models are
restricted to finite ones. To lift the restriction, some numerical
measures, such as the probability, should be added to the models.

\subsection{Arbitration}
The notion of distance measure between possible worlds is also
used in another type of merging operator, called
arbitration\cite{lib1,rev2,rev1}. Arbitration is the process of
settling a conflict between two or more persons. The first version
of arbitration operator between knowledge bases is proposed in
\cite{rev2} via the so-called model-fitting operators. The
postulates for model-fitting operators and its semantic
characterization  are given and then arbitration is defined as a
special kind of model-fitting operators.

In \cite{rev1}, the arbitration operator is further generalized so
that it is applicable to the weighted knowledge bases. A set of
postulates is also directly used in  characterizing  the
arbitration between a weighted knowledge base and a regular one. A
weighted knowledge base in \cite{rev1} is defined as a mapping
$\tilde{K}$ from model sets to nonnegative real number and a
regular knowledge base is just a finite set of propositional
sentences. A generalized loyal assignment is then defined as a
function that assigns for each weighted knowledge base $\tilde{K}$
a pre-order $\leq_{\tilde{K}}$ between propositional sentences
such that some conditions are satisfied for the pre-orders.
Finally, the arbitration of a weighted knowledge base $\tilde{K}$
by a regular knowledge base $K'$ is defined as
\[\tilde{K}\triangle K'=\min(K',\leq_{\tilde{K}}),\]
where $\min(K',\leq_{\tilde{K}})$ is the set of sentences in $K'$
which is minimal according to the ordering $\leq_{\tilde{K}}$.
However, this kind of arbitration is obviously syntax-dependent.
For example, if $\varphi_1$ and $\varphi_2$ is two propositional
sentences such that $\varphi_1<_{\tilde{K}}\varphi_2$, then
$\tilde{K}\triangle
\{\varphi_1,\varphi_2\}=\{\varphi_1\}\not=\tilde{K}\triangle
\{\varphi_1\wedge\varphi_2\}=\{\varphi_1\wedge\varphi_2\}$ though
the two knowledge bases $\{\varphi_1, \varphi_2\}$ and
$\{\varphi_1\wedge\varphi_2\}$ are semantically equivalent.

An alternative, seemingly more natural, characterization for
arbitration is given in \cite{lib1} without resorting to the
model-fitting operators. A  knowledge base in that work is
identified with the set of propositional models for it, thus the
semantic characterization for this kind of arbitration is given by
assigning to each subset of models $A$ a binary relation $\leq_A$
over the set of model sets satisfying the following conditions
(the subscript is omitted when it means all binary relations of
the form $\leq_A$)
\begin{enumerate}
\item transitivity: if $A\leq B$ and $B\leq C$ then $A\leq C$
\item if $A\subseteq B$ then $B\leq A$
\item $A\leq A\cup B$ and $B\leq A\cup B$
\item $B\leq_A C$ for every $C$ iff $A\cap B\not=\emptyset$
\item $A\leq_{C\cup D}B\Leftrightarrow\left\{\begin{array}{ll}
 C\leq_{A\cup B} D \;\; and \;\; A\leq _C B \;\; or\\
D\leq_{A\cup B} C \;\; and \;\; A\leq _D B
\end{array}
\right.$
\end{enumerate}
Then the arbitration between two sets of models $A$ and $B$ is
defined as
\begin{equation}\label{eq10}
A\triangle B=\min(A,\leq_B)\cup\min(B,\leq_A)
\end{equation}
Note that though the relation $\leq_A$ is defined between sets of
models, in the definition of the arbitration, only $\leq_A$
between singletons is used. Thus by slightly abusing the notation,
$\leq_A$ may also denote an ordering between models.

To incorporate the arbitration operator of \cite{lib1} into our
langauge, we must first note that according to (\ref{eq10}), the
arbitration is commutative but not necessarily associative. Thus,
the arbitration operator should be a binary one between two
agents. We can add a class of modal operators for arbitration into
our logic just as in the case of majority merging. However, to be
more expressive, we will also consider the interaction between
arbitration and other epistemic operators, so we define the set of
{\em arbitration expressions \/} over the agents recursively as
the smallest set containing $\{1,2,\cdots,n\}$ and closed under
the binary operators $+, \cdot$, and $\triangle$. Here $+$ and
$\cdot$ correspond respectively to the distributed belief and the
so-called ``everybody knows'' operators in multi-agent epistemic
logic\cite{fag}. Then our language can be extended to include a
new class of modal operators $[a]$ where $a$ is an arbitration
expressions. Note that it has been shown that the only associative
arbitration satisfying postulates 7 and 8 of \cite{lib1} is
$A\triangle B=A\cup B$, so if $\triangle$ is an associative
arbitration satisfying those postulates, then $[a\triangle
b]\varphi$ is reduced to $[a\cdot b]\varphi$ which is in turn
equivalent to $[a]\varphi\wedge[b]\varphi$.

For the semantics, a model is extended to $(W, ({\cal R}_i)_{1\leq
i\leq n}, V, \leq)$, where $(W, ({\cal R}_i)_{1\leq i\leq n}, V)$
is a DBF$_n^s$ (or DBF$_n^c$) model, whereas $\leq$ is a function
assigning to each subset of possible worlds $A$ a binary relation
$\leq_A\subseteq 2^W\times 2^W$ satisfying the above-mentioned
five conditions. Note that the first two conditions imply that
$\leq_A$ is a pre-order over $2^W$. Then for each arbitration
expression, we can define the binary relations ${\cal
R}_{a\triangle b}, {\cal R}_{a\cdot b}$ and ${\cal R}_{a+b}$ over
$W$ recursively by
\begin{equation}\label{eq11}{\cal
R}_{a\triangle b}(w)=\min({\cal R}_a(w),\leq_{{\cal
R}_b(w)})\cup\min({\cal R}_b(w),\leq_{{\cal
R}_a(w)})\end{equation}
\begin{equation}\label{eq12}{\cal R}_{a+b}={\cal R}_a\cap {\cal R}_b\end{equation}
\begin{equation}\label{eq13}{\cal R}_{a\cdot b}={\cal R}_a\cup {\cal R}_b\end{equation}
Thus the satisfaction for the wff $[a]\varphi$ is defined as
\[u\models[a]\varphi \;\; \mbox{\rm iff for all}
\;\;w\in {\cal R}_a(u), w\models\varphi.\] Note that the
distributed belief operator $[G]$ can be equivalently defined as
an abbreviation of $[i_1+(i_2+\cdots(i_{k-1}+i_k))]$ if
$G=\{i_1,i_2,\cdots,i_k\}$.

By this kind of modal operators, the postulates 2-8 of \cite{lib1}
can be translated into the following axioms:
\begin{enumerate}
\item $[a\triangle b]\varphi\equiv[b\triangle a]\varphi$
\item $[a\triangle b]\varphi\supset[a+b]\varphi$
\item $\neg[a+b]\bot\supset([a+b]\varphi\supset[a\triangle
b]\varphi)$
\item $[a\triangle b]\bot\supset[a]\bot\wedge[b]\bot$
\item $([a\triangle(b\cdot c)]\varphi\equiv[a\triangle
b]\varphi)\vee ([a\triangle(b\cdot c)]\varphi\equiv[a\triangle
c]\varphi)\vee ([a\triangle(b\cdot c)]\varphi\equiv[(a\triangle
b)\cdot(a\triangle c)]\varphi)$
\item $[a]\varphi\wedge[b]\varphi\supset[a\triangle b]\varphi$
\item $\neg[a]\bot\supset\neg[ a+(a\triangle b)]\bot$
\end{enumerate}
However, since the set of possible worlds $W$ may be infinite in
our logic, the minimal models in (\ref{eq11}) may not exist, so
the axioms 4 and 7 are not sound with respect to the semantics. To
make them sound, we must add the following limit
assumption\cite{arlo} to the binary relations $\leq_A$ for any
$A\subseteq W$:
\begin{center}
  for any nonempty $U\subseteq W$, $\min(U,\leq_A)$ is nonempty.
\end{center}
It must be noted that the axioms listed above are not yet complete
for the logic with arbitration operators. In fact, the search of a
complete axiomatization for the modal logic of arbitration
expressions is of independent interest by itself  and can be the
further research direction. The brief presentation here just shows
that the modal logic approach  can provides a uniform framework
for integrating the epistemic reasoning and different knowledge
merging operators into the object logic level.

\subsection{General merging}\label{sec:merg}
In \cite{kon1}, an axiomatic framework unifying the majority
merging  and arbitration operators is presented. A set of
postulates common to majority and arbitration operators is first
proposed to characterize the general merging operators and then
additional postulates for differentiating these two are considered
respectively. In the framework, a knowledge base is also a finite
set of propositional sentences. The general merging operator is
defined as a mapping from a multi-set(also called a
bag)\footnote{A multi-set (or bag) is a collection of elements
over some domain which allows multiple occurrences of elements.}
of knowledge base (called a knowledge set) to a knowledge base.
Thus the arbitration operator defined via the approach can merge
more than two knowledge bases whereas the arbitration operator in
\cite{lib1} is defined only for two knowledge bases. The merging
operator is denoted by $\triangle$, so for each knowledge set $E$,
$\triangle(E)$ is a knowledge base. Two equivalent semantic
characterizations are also given for the merging operators. One of
them is based on the so-called syncretic assignment. A syncretic
assignment maps each knowledge set $E$ to a pre-order $\leq_E$
over interpretations such that some conditions reflecting the
postulated properties of the merging operators must be satisfied.
Then $\triangle(E)$ is the knowledge base whose models are the
minimal interpretations according to $\leq_E$.

This logical framework is further extended to dealing with
integrity constraints in \cite{kon2}. Let $E$ be a knowledge set
and $\varphi$ is a propositional sentence denoting the integrity
constraints, then the merging of knowledge bases in $E$ with
integrity constraint $\varphi$, $\triangle_\varphi(E)$, is a
knowledge base which implies $\varphi$. The models of
$\triangle_\varphi(E)$ is characterized by
$\min(Mod(\varphi),\leq_E)$, i.e., the minimal models of $\varphi$
with respect to the ordering $\leq_E$. $\triangle_\varphi(E)$ is
called IC merging operator. According to  the semantics, it is
obvious that $\triangle(E)$ is a special case of IC merging
operator $\triangle_\top(E)$. It is also shown that when $E$
contains exactly one knowledge base, the operator is reduced to
the AGM revision operator proposed in \cite{agm}. Thus IC merging
is general enough to covering the majority merging, arbitration
and AGM revision operator.

To realize the IC merging operators in the modal logic framework,
we will extend the syntax of our logic with the following
formation rule:
\begin{itemize}
\item if $\varphi$ and $\psi$ are wffs, then for any nonempty
$G\subseteq\{1,2,\ldots,n\}$, $[\triangle_\varphi(G)]\psi$ is also
a wff.
\end{itemize}
For the naming convenience, we will call a subset of possible
worlds a belief state. Let ${\cal U}=\{U_1,U_2,\ldots,U_k\}$
denote a multi-set of belief states, then $\bigcap{\cal
U}=U_1\cap\cdots U_k$. For the semantics, a possible world model
is extended to $(W, ({\cal R}_i)_{1\leq i\leq n}, V, \leq)$, where
$(W, ({\cal R}_i)_{1\leq i\leq n}, V)$ is a DBF$_n^s$ (or
DBF$_n^c$) model, whereas $\leq$ is an assignment mapping each
multi-set of belief states ${\cal U}$ to a total pre-order
$\leq_{\cal U}$ over $W$ satisfying the following conditions:
\begin{enumerate}
\item If $w, w'\in\bigcap{\cal U}$, then $w\leq_{\cal U}w'$
\item If $w\in\bigcap{\cal U}$ and  $w'\not\in\bigcap{\cal U}$
then $w<_{\cal U}w'$
\item For any $w\in U_1$, there exists $w'\in U_2$, such that
$w'\leq_{\{U_1,U_2\}} w$, where $U_1$ and $U_2$ are two belief
states
\item If $w\leq_{{\cal U}_1}w'$ and $w\leq_{{\cal U}_2}w'$, then
$w\leq_{{\cal U}_1\sqcup{\cal U}_2}w'$, where $\sqcup$ denotes the
union of two multi-sets
\item If $w<_{{\cal U}_1}w'$ and $w\leq_{{\cal U}_2}w'$, then
$w<_{{\cal U}_1\sqcup{\cal U}_2}w'$
\end{enumerate}
For a sub-group of agents $G$ and a possible world $u$, let us
define a total pre-order $\leq_G^u$ over $W$ as follows:
\[w\leq_G^u w'\;\;{\rm iff}\;\; w\leq_{\{{\cal R}_i(u)\mid i\in G\}}w'.\]
Then a possible world $u$ satisfies the wff
$[\triangle_\varphi(G)]\psi$ in the model, i.e.\
$u\models[\triangle_\varphi(G)]\psi$, iff
\begin{description}
\item (i) there are no possible worlds in $W$ satisfying $\varphi$, or
\item (ii) there exists $w_0\in W$ such that $w_0\models\varphi$ and for
any $w\leq_G^u w_0$, $w\models\varphi\supset\psi$.
\end{description}

Though we can incorporate the general merging operator into the
modal logic framework, we should not overlook the difference
between the meta-level merging operators and the modal ones.
First, in the meta-level approach, the knowledge set consists of a
multi-set of objective sentences, whereas in the modal operator
$[\triangle_\varphi(G)]$, $G$ is a set of agents whose belief may
contains subjective sentences or beliefs of other agents. Second,
the integrity constraint can only be the objective sentences in
\cite{kon2} whereas $\varphi$ may be arbitrary complex wff of our
extended language. Finally, instead of selecting the minimal
models from those of $\varphi$, since the set of possible worlds
may be infinite in our case, we adopt the system-of-spheres
semantics as that in section \ref{sec:major} for the modal
operator $[\triangle_\varphi(G)]$.

\section{Belief Change and Conditional Logic}\label{sec7}
\subsection{Incorporating belief revision operators}
Unlike knowledge merging, where the component knowledge bases are
equally important, belief change is a kind of asymmetry operators,
where the new information always outweighs the old one. The main
belief change operators are belief revision and update. They are
characterized by different postulates\cite{agm,kat1,kat2}. In
\cite{kat1}, a uniform model-theoretic framework is provided for
the semantic characterization of the revision and update
operators. In their works, a knowledge base is a finite set of
propositional sentences, so it can also be represented by a single
sentence(i.e., the conjunction of all sentences in the knowledge
base).

For the revision operator, it is assumed that there is a total
pre-order $\leq_\psi$ over the propositional interpretations for
each knowledge base $\psi$. The revision operators satisfying the
AGM postulates in \cite{agm} are exactly those that select from
the models of the new information $\varphi$ the minimal ones with
respect to the ordering $\leq_\psi$. More precisely, let $\psi$ be
a knowledge base and $\varphi$ denote the new information, then
the result of revising $\psi$ by $\varphi$, denoted by
$\psi\circ\varphi$, will have the set of models
\[Mod(\psi\circ\varphi)=\min(Mod(\varphi),\leq_\psi).\]

As for the update operator, assume for each propositional
interpretation $w$, there exists some partial pre-order $\leq_w$
over the interpretations for closeness to $w$, then update
operators select for each model $w$ in $Mod(\psi)$ the set of
models from $Mod(\varphi)$ that are closest to $w$. The updated
theory is characterized by the union of all such models. That is,
\[Mod(\psi\diamond\varphi)=\bigcup_{w\in
Mod(\psi)}\min(Mod(\varphi),\leq_w)\] where $\psi\diamond\varphi$
is the result of updating the knowledge base $\psi$ by $\varphi$.

Both belief revision and update may occur in the observation of
the new information $\varphi$. For the belief revision, it is
assumed that the world is static, so if the new information is
incompatible with the agent's original beliefs, then the agent may
have wrong belief about the world. Thus he will try to accommodate
the new information by minimally changing his original beliefs.
However, for the belief update, it is assumed that the observation
may be due to the dynamic change of the outside world, so the
agent's belief may be out-of-date, though it may be totally
correct for the original world. Thus the agent will assume the
possible worlds are those resulting from the minimal change of the
original world. In \cite{bou5}, a generalized update model is
proposed which combines aspects of revision and update. It is
shown that a belief update model will be inadequate without
modelling the dynamic aspect (i.e.\ the events causing the update)
in the same time. Since the dynamic change of the external worlds
does not play a role in the belief fusion process, we would not
try to model the belief update in our logic, so in what follows,
we will concentrate on the belief revision operator.

Let us now consider the possibility of incorporating the belief
revision operator into our logic. In addition to the original
meaning of revising a knowledge base $\psi$ by new information
$\varphi$, there is an alternative reading for the revision
operator. That is, we can  consider $\circ$ as a prioritized
belief fusion operator  which gives the priority to its second
argument\cite{sho2}. In the context of knowledge base revision,
these two interpretations are essentially equivalent. However,
from the perspective of our logic in multi-agents systems, they
may be quite different. Roughly speaking,  $i\circ\varphi$ will
denote the result of revising the beliefs of agent $i$ by new
information $\varphi$, whereas $i\circ j$ is the result of merging
the beliefs of agents $i$ and $j$ by giving priority to $j$. More
formally, an revision expression will be defined inductively as
follows:
\begin{itemize}
\item If  $1\leq i, j\leq n$ and $\varphi$ is a wff, then $i\circ j$ and $i\circ\varphi$ are
revision expressions.
\item If $r$ is a revision expression, $1\leq i\leq n$  and $\varphi$ is a wff, then $r\circ i$
and $r\circ\varphi$ are revision expressions.
\end{itemize}
The syntactic rule is extended to include the modal operators
$[r]$ for any revision expression $r$, so $[r]\varphi$ would be a
wff if $\varphi$ is. To interpret the modal operator in our
semantic framework,  a possible world model is extended to $(W,
({\cal R}_i)_{1\leq i\leq n}, V, \leq)$, where $(W, ({\cal
R}_i)_{1\leq i\leq n}, V)$ is a DBF$_n^s$ (or DBF$_n^c$) model,
whereas $\leq$ is an assignment mapping each belief state (i.e.\
subset of possible worlds) $U$ to a total pre-order $\leq_U$ over
$W$ such that (i) if $w, w'\in U$, then $w\leq_U w'$ and (ii)~if
$w\in U$ and $w'\not\in U$, then $w<_U w'$.  Let $S\cdot U$ denote
the sequence $(U_1,U_2,\cdots, U_k, U)$ if $S=(U_1,U_2,\cdots,
U_k)$ is a sequence of belief state, then the assignment $\leq$ is
extended to sequences of belief states in the following way (we
assume $\leq_{(U)}=\leq_U$):
\begin{enumerate}
\item $w<_{S\cdot U}w'$ if $w\in U$ and $w'\not\in U$
\item $w\leq_{S\cdot U}w'$ iff $w\leq_S w'$ when $w,w'\in U$ or $w,w'\not\in U$
\end{enumerate}
For each wff $\varphi$, let the truth set of $\varphi$, denoted by
$|\varphi|$, be defined as $\{w\in W\mid w\models\varphi\}$. For
each possible world $u$, define a function mapping any agent $i$
and revision expression $r$ into a sequence of belief states
$u(i)$ and $u(r)$ as follows:
\begin{enumerate}
\item $u(i)=({\cal R}_i(u))$
\item $u(r\circ i)=u(r)\cdot{\cal R}_i(u)$
\item $u(r\circ\varphi)=u(r)\cdot|\varphi|$
\end{enumerate}
Then the truth condition for the wff $[r\circ\varphi]\psi$ is
$u\models[r\circ\varphi]\psi$ iff
\begin{description}
\item (i) there are no possible worlds in $W$ satisfying $\varphi$, or
\item (ii) there exists $w_0\in W$ such that $w_0\models\varphi$ and for
any $w\leq_{u(r)} w_0$, $w\models\varphi\supset\psi$.
\end{description}
Analogously,  $u\models[r\circ i]\psi$ iff there exists  $w_0\in
{\cal R}_i(u)$ such that for any $w\leq_{u(r)} w_0$, if $w\in
{\cal R}_i(u)$, then $w\models\psi$. It can be seen that
$[i\circ\varphi]\psi$ is equivalent to
$[\triangle_\varphi(\{i\})]\psi$ in section \ref{sec:merg}
according to the semantics.

\subsection{Relationship with conditional logic}
There have been also various attempts in formalizing the belief
change process by modal logic or conditional logic
systems\cite{bou1,bou2,bou3,bou4,nir,ryan,seg}. For example, in
\cite{bou2}, a modal logic CO$^*$ is proposed for modelling the
belief revision. CO$^*$ is an extension of the logic CO proposed
in \cite{bou1}. In CO$^*$, revision of a theory by a sentence is
represented using a conditional connective. The connective is not
primitive, but rather defined using two unary modal operators
$\Box$ and $\stackrel{\leftarrow}{\Box}$. The modal operators are
interpreted with respect to a total pre-order $R$ over the
possible worlds which is assumed to rely on a background theory
$K$. Thus $w\models\Box\varphi$ iff $\varphi$ is true in all
possible worlds which are as plausible as $w$ given the theory $K$
and $w\models\stackrel{\leftarrow}{\Box}\varphi$ iff $\varphi$ is
true in all possible worlds which are less plausible than $w$
given $K$. By defining $\stackrel{\leftrightarrow}{\Box}\varphi$
as $\Box\varphi\wedge\stackrel{\leftarrow}{\Box}\varphi$ and
$\stackrel{\leftrightarrow}{\Diamond}\varphi$ as
$\neg\stackrel{\leftrightarrow}{\Box}\neg\varphi$, the conditional
$\varphi\stackrel{KB}{\ra}\psi$ is defined as
\[\stackrel{\leftrightarrow}{\Box}\neg\varphi\vee
\stackrel{\leftrightarrow}{\Diamond}(\varphi\wedge\Box(\varphi\supset\psi)),\]
where $KB$ is a finite representation of the theory $K$. Since
there is only one global ordering $R$ in CO$^*$ model which is
associated with the background theory, the logic is appropriate
only for reasoning about the revision of a single theory $K$. On
the other hand, our logic allows the reasoning about revisions of
many agents' belief states. Furthermore, since the ordering $R$ in
CO$^*$ model is global, $\varphi\stackrel{KB}{\ra}\psi$ is true in
a world iff it is true in all worlds, thus no iterated revisions
are allowed in the model. In \cite{bou4}, this restriction is
lifted by allowing the revision of the ordering $R$ to $R'$ at the
same time. The idea is to move the most plausible $\varphi$-models
with respect to $R$ to the most plausible level of $R'$ and keep
the other parts of $R$ unchanged. Our assignment of a total
pre-order to a sequence of belief states is basically based on the
same idea. However, while the definition of \cite{bou4} presumes
the existence of the minimal models for any propositional
formulas, we do not need such assumption.

In \cite{nir}, a logic with conditional and epistemic operators is
used in the reasoning of belief revision. The conditional and
epistemic sentences are interpreted in an abstract belief change
system (BCS). The basic components of a BCS are a set of belief
states and a belief change function mapping each belief state and
sentence of some base language into  a new belief state. In a more
concrete preferential interpretation, each belief state $s$  is
interpreted as a subset of possible world $K(s)$ and a pre-order
$\preceq_s$ over the possible worlds is associated with it. In
this regard, $\preceq_s$ corresponds to $\leq_{K(s)}$ in our
semantic models and the conditional wff $\varphi>\psi$ in the
logic ${\cal L}^>$ of \cite{nir} is roughly equivalent to our wff
$[i\circ\varphi]\psi$ for some fixed agent $i$. However, since in
${\cal L}^>$, the antecedent $\varphi$ of a conditional is
restricted to a wff in the base language $\cal L$, it does not
allow the epistemic wffs of the form $B\varphi$. It is argued that
the antecedent must be observable whereas conditional wffs are
unobservable, so we should not  allow conditional wffs in the
antecedent. However, from the multi-agent systems perspective, one
agent may learn the beliefs of other agents by communication, so
we should not exclude the flexibility. Another significant
difference is that the BCS allows only revision of a belief state
by a sentence, while in our system, the prioritized fusion of two
belief states held by two agents are also incorporated.

A dynamic doxastic logic for belief revision is proposed in
\cite{seg} and further developed in \cite{seg1}. By using the
notations of \cite{seg}, the doxastic operator $B$ and two kinds
of dynamic modal operators $[+\varphi]$ and $[-\varphi]$ for
propositional wff $\varphi$ are taken as the basic constructs of
the language. The operators $[+\varphi]$ and $[-\varphi]$
corresponds respectively to the expansion and contraction
operators in AGM theory. Thus the revision operator
$[\circ\varphi]$ is defined as $[-(\neg\varphi)][+\varphi]$
according to the so-called Levi's identity\cite{agm}. The wffs of
the language are interpreted with respect to  a hypertheory. A
hypertheory $H$ is a set of subsets of possible worlds linearly
ordered by inclusion. A hypertheory is similar to the widening
ranked model defined in \cite{lehm}. However, the latter assumes
that the subsets of models are indexed by natural numbers. A
hypertheory is said to be replete if the set of all possible
worlds $W$ is in $H$. From the hypertheory $H$, a pre-order
$\leq_H$ over $W$ can be defined as follows:
\begin{center}
$w'\leq_H w$ iff for any $U\in H$, if $w\in U$, then there exists
$U'\in H$ such that $U'\subseteq U$ and $w'\in U'$.
\end{center}
When $H$ is replete, the pre-order defined in this way is total.
When the wffs $[+\varphi]\psi$ and $[-\varphi]\psi$ are evaluated
with respect to a hypertheory $H$, it causes evaluation of $\psi$
in some revised hypertheory $H'$, so the semantics are essentially
equivalent to that proposed in \cite{bou4}, though the revisions
of the corresponding pre-order are somewhat different in the two
approaches. Therefore, as the proposal of \cite{bou4}, the logic
allows only reasoning about the belief revision of a single agent
by some new information and the prioritized fusion of multi-agent
beliefs can not be represented in such logic.

\subsection{Alternative representations of belief states}
In our presentation above, we assume an agent's belief states are
represented as a subset of possible worlds, i.e.\ ${\cal R}_i(w)$
is the belief state of agent $i$ in world $w$. However, some more
fine-grained representations have been also proposed, such as
total pre-orders over the set of possible worlds
\cite{bou4,dar,lehm,seg}, ordinal conditional functions
\cite{bou5,spo,will}, possibility
distributions\cite{ben2,dp5,dp4}, belief functions\cite{smet} and
pedigreed belief states\cite{sho2}. Perhaps, the most popular
representation among them is an ordering of the possible worlds.
While a set of possible worlds can be seen as the minimal worlds
with respect to  a given ordering, it is claimed that the fusion
of two orderings is more general than the revision of an ordering
by a set of possible worlds\cite{sho2}. Thus it is shown that AGM
revision is in fact a special case of the fusion operator in
\cite{sho2}. Indeed, in our extended models, the assignment $\leq$
has mapped each subset of possible worlds to a total pre-orders
between worlds. However, to fully utilize the semantic power of an
ordering, the logic language should be extended further to cover
the conditional connectives. Since the purpose of the present
paper is to integrate the belief fusion operators into the
epistemic reasoning framework, this extension is beyond its scope.
Nevertheless, the further development of logical systems
incorporating the fusion operators based on fine-grained
representations of belief states should be a very interesting
research direction.

\section{Conclusions and Further Researches}
The main contribution of the paper is the integration of belief
fusion operators into the multi-agent epistemic logic. We first
propose two basic logical systems for reasoning about the
cautiously merged beliefs of multiple agents and then extend them
to cover more sophisticated and adventurous fusion and revision
operators.

The basic systems are cautious in the sense that if an information
source is in conflict with other more reliable ones, then the
information from that source is totally discarded. The two systems
correspond to two different strategies of discarding the
information sources. For level cutting strategy, if an information
source is to be discarded, then all those less reliable than it
are also discarded without further examination. On the other hand,
for level skipping strategy, only the level under conflict is
skipped, and the next level will be considered independent of
those discarded before it. Thus, level skipping strategy is
relatively less cautious than the level cutting one and indeed, we
can simulate the trusting attitude multi-sources reasoning in
\cite{cho} by level skipping strategy.

Then some of the most important knowledge base merging approaches
are reviewed and it is shown that many fusion operators proposed
in those approaches can be incorporated into our logic. While most
of the knowledge base merging research takes the fusion process as
a meta-level operator, our approach incorporate them into the
object logic directly. Therefore, it is possible to integrate the
belief fusion operators into the multi-agent epistemic logic. What
we can benefit from the epistemic logic is the capability to
reasoning with not only the beliefs about the objective world but
also the beliefs about beliefs.

In the discussion of the belief fusion logic, for simplicity, we
do not distinguish belief and information. However, in a genuine
agent systems, an agent's belief may be different than the
information he sends to or receives from other agents. Thus, in
general, we should have  a set of modal operators $[j]_i$ such
that $[j]_i\varphi$ means that agent $i$ receives the information
$\varphi$ from $j$. In particular, $[i]_i\varphi$ may represent
the observation of agent $i$ himself, which should be the most
reliable information source for $i$. Then agent $i$ may form his
belief by fusing the information he received from different agents
according to the degrees of trust he has on other agents. The
fusion may be represented by the operators $[O]_i$. If we consider
$[j]_i\varphi$ as the communication of message $\varphi$ from $j$
to $i$, then we have a general framework for reasoning about
agent's belief and communication. In such a framework, we can
discuss the problems like deception of agent. For example,
$[O]_i\varphi\wedge[i]_j\neg\varphi$ may mean that agent $i$
deceives to agent $j$ by telling $j$ the negation of what he
believes. In \cite{liau2}, an application of our basic systems to
reasoning about beliefs and trusts of multiple agents has been
proposed along this direction. However, more works remain to be
done for the real applications. These applications may also
benefit from some fundamental works on multi-agent belief
revision\cite{drag1,drag2,drag3,gall,tenn}.

From a more foundational viewpoint, though we have proposed some
extensions of the basic systems for accommodating the belief
fusion operators both syntactically and semantically, the complete
axiomatization of these extended logics remain to be found. To be
practically useful, other proof methods more suitable for
automated theorem proving should be also developed.

In a recent paper, it is shown that the multi-sources reasoning
can be applied to deontic logic under conflicting
regulations\cite{cho1}. Essentially, this is to merge conflicting
regulations according to the priorities of them analogously to the
fusion of information. However, inherited from the restriction of
FU$_n$, it is also required that each regulation to be merged must
be a set of deontic literals. Now, by the systems developed here,
it is expected that the general forms of regulations can also be
merged.

A real difficulty in the application of our logic to model the
database merging reasoning is the representational problem of the
databases. In the discussion of section \ref{sec5}, we suggest to
find all maximal consistent agent groups in advance and add the
wff $\bigwedge_{G\in MCAG}\neg[G]\bot$ to the representation. This
is a rather time-consuming work. In practice, we can omit this
part and check the consistency of some agent groups when it is
necessary during the course of proof. Even further, we can
consider the implementation of the logic with some non-monotonic
reasoning techniques\cite{anto} so that only the explicit
information in the databases have to be represented by the wffs.
This will be investigated in the further research.

\bibliography{bibfile}

\appendix
\section{Proof of the Proposition and Theorems}
\subsection{Proof of Proposition \ref{prop1}}
\begin{enumerate}
\item By induction on $m=|\delta(O)|$:
\begin{description}
\item $m=1$: this is trivial since we identify $[i_1]\varphi$ and
$[\{i_1\}]\varphi$ in our language.
\item assume this result holds for all $m\leq k$.
\item $m=k+1$: there are two cases:
\begin{description}
\item $j=m$:
$\vdash\neg[G_j]\varphi\supset([O]\varphi\equiv[G_j]\varphi)$ is
just an instance of axiom O1,
\item $j<m$: let $O$ be written as $O'>i_m$ where
$O'=i_1>\ldots>i_k$, then this proof is as follows:
\[\begin{array}{ll}
1. \neg[G_j]\bot\wedge[G_{j+1}]\bot\supset[G_m]\bot & G3, m\geq
j+1\\ 2. [G_m]\bot\supset([O]\varphi\equiv[O']\varphi) & O2\\ 3.
\neg[G_j]\bot\wedge[G_{j+1}]\bot\supset
([O]\varphi\equiv[O']\varphi) & 1,2,P, MP\\ 4.
\neg[G_j]\bot\wedge[G_{j+1}]\bot\supset
([O']\varphi\equiv[G_j]\varphi) & \mbox{\rm ind. hyp.}\\ 5.
\neg[G_j]\bot\wedge[G_{j+1}]\bot\supset([O]\varphi\equiv[G_j]\varphi)
& 3, 4, P, MP
\end{array}
\]
\end{description}
\end{description}
\item By induction on $|\delta(O)|$: if $|\delta(O)|=1$, this is
an instance of G1. Assume this holds for modal operator $[O]$, let
us consider the proof for $[O>i]$. Let $p$ and $G$ denote
respectively $[\delta(O>i)]\bot$ and $\delta(O>i)$
\[\begin{array}{ll}
1. [O>i]\varphi\supset(\neg p\supset[G]\varphi) & O1\\ 2.
[O>i](\varphi\supset\psi)\supset(\neg
p\supset[G](\varphi\supset\psi)) & O1\\ 3.
[O>i]\varphi\supset(p\supset[O]\varphi) & O2\\ 4.
[O>i](\varphi\supset\psi)\supset(p\supset[O](\varphi\supset\psi))
& O2\\ 5. [O>i]\varphi\wedge[O>i](\varphi\supset\psi)\supset(\neg
p\supset[G]\psi) &1, 2, G1, P, MP\\ 6.
[O>i]\varphi\wedge[O>i](\varphi\supset\psi)\supset(
p\supset[O]\psi) &3, 4, \mbox{ \rm ind. hyp., } P, MP\\ 7.
[O>i]\varphi\wedge[O>i](\varphi\supset\psi)\supset(\neg
p\supset[O>i]\psi) &5, O1, P, MP\\ 8.
[O>i]\varphi\wedge[O>i](\varphi\supset\psi)\supset(
p\supset[O>i]\psi) &6, O2, P, MP\\ 9.
[O>i]\varphi\wedge[O>i](\varphi\supset\psi)\supset[O>i]\psi &7, 8,
P, MP
\end{array}
\]
\item By induction on $|\delta(O)|$: if $|\delta(O)|=1$, this is
an instance of G2. Assume we have $\vdash\neg[O]\bot$, then the
proof of $\vdash\neg[O>i]\bot$ is as follows:
\[\begin{array}{ll}
1. \neg[\delta(O>i)]\bot\supset([O>i]\bot\supset[\delta(O>i)]\bot)
& O1\\ 2. [\delta(O>i)]\bot\supset([O>i]\bot\supset[O]\bot) & O2\\
3. [O>i]\bot\supset[\delta(O>i)]\bot & 1, P, MP\\ 4.
[O>i]\bot\supset([\delta(O>i)]\bot\supset[O]\bot) & 2, P, MP\\ 5.
[O>i]\bot\supset[O]\bot & 3, 4, P, MP\\ 6. \neg[O]\bot & \mbox{\rm
ind. hyp.}\\ 7. \neg[O>i]\bot & 5, 6, P, MP
\end{array}
\]
\item By induction on $|\delta(O)|$: if $|\delta(O)|=1$, this is
an instance of Gen rule. Assume it is the case for modal operator
$[O]$, let us consider the proof for $[O>i]$
\[\begin{array}{ll}
1. \varphi & \mbox{\rm Assumption}\\ 2. [O]\varphi & \mbox{\rm
Ind. Hyp.}\\ 3. [\delta(O>i)]\varphi & Gen\\ 4.
[O>i]\varphi\equiv([\delta(O>i)]\varphi\vee[O]\varphi) & O1, O2,
P, MP\\ 5. [O>i]\varphi & 2, 3, 4, P, MP
\end{array}
\]
\end{enumerate}
$\Box$

\subsection{Proof of Theorem \ref{thm1}}
The proof of the theorem is based on that for S5$_n^D$ in
\cite{fag,fag1}. As usual, the verification of the soundness part
is a routine checking, so we focus on the completeness part. Let
$\cal L$ denote a logical system.  A wff $\varphi$ is $\cal
L$-inconsistent if its negation $\neg\varphi$ can be proved in
$\cal L$. Otherwise, $\varphi$ is $\cal L$-consistent. A set
$\Sigma$ of wffs is said to be $\cal L$-inconsistent if there is a
finite subset $\{\varphi_1,\ldots, \varphi_k\}\subseteq\Sigma$
such that the wff $\varphi_1\wedge\cdots\wedge\varphi_k$ is $\cal
L$-inconsistent; otherwise, $\Sigma$ is $\cal L$-consistent. A
maximal $\cal L$-consistent set of wffs ($\cal L$-MCS) is a
consistent set $\chi$ of wffs such that whenever $\psi$ is a wff
not in $\chi$, then $\chi\cup\{\psi\}$ is $\cal L$-inconsistent.

On the other hand, $\varphi$ is $\cal L$-satisfiable iff there
exists a $\cal L$ model $M$ and a possible world $w$ such that
$w\models_M\varphi$, otherwise $\varphi$ is $\cal
L$-unsatisfiable. Sometimes the prefix $\cal L$ will be omitted
without confusion. To prove the completeness, we will show that
every DBF$_n^c$-consistent wff is DBF$_n^c$-satisfiable.

Let ${\cal I}={\cal TO}_n\cup 2^{\{1,2\ldots,n\}}-\{\emptyset\}$
be the set of all modal operators for the language DBF$_n^c$. A
{\em pseudo \/} DBF$_n^c$ structure is a tuple $(W,({\cal
R}^*_I)_{I\in{\cal I}}, V)$ where $W$ and $V$ are defined as in
DBF$_n^c$ models and each ${\cal R}^*_I$ is a binary relation on
$W$. Furthermore,it is required that ${\cal R}^*_{\{i\}}$ is a
serial relation for each $1\leq i\leq n$. The satisfaction clauses
for DBF$_n^c$ wffs in pseudo structures are defined as usual, so
for example, we have $w\models[O]\varphi$ iff for $u\in{\cal
R}^*_O(w)$, $u\models\varphi$. What make difference is that in a
pseudo structures, each ${\cal R}^*_I$ is considered as an
independent relation instead of the intersection of other
individual ones. A pseudo structure $M^*$ is called a pseudo model
if all  wffs provable in  DBF$_n^c$ are valid in $M^*$. A
DBF$_n^c$ wff $\varphi$ is pseudo satisfiable if there exists a
pseudo model $M^*$ and a possible world $w$ such that
$w\models_{M^*}\varphi$.

The following two results will be proved:
\begin{mylma}\label{lma1}
\begin{enumerate}
\item If $\varphi$ is DBF$_n^c$-consistent, then $\varphi$ is pseudo
DBF$_n^c$-satisfiable.
\item If $\varphi$ is pseudo DBF$_n^c$-satisfiable, then it is DBF$_n^c$-satisfiable.
\end{enumerate}
\end{mylma}

The first result is proved by a standard canonical model
construction procedure. A canonical pseudo structure
$M^*=(W,({\cal R}_I)^*_{I\in{\cal I}}, V)$ is defined as follows
\begin{itemize}
\item $W=\{w_\chi\mid\chi$ is a DBF$_n^c$-MCS$\}$, in other words, each possible world corresponds
precisely to  a DBF$_n^c$-MCS.
\item ${\cal R}^*_I(w_{\chi_1},w_{\chi_2})$ iff
$\chi_1/I\subseteq\chi_2$ for all $I\in\cal I$, where
$\chi_1/I=\{\varphi\mid[I]\varphi\in\chi_1\}$.
\item $V:\Phi_0\ra 2^W$ is defined by $V(p)=\{w_\chi\mid
p\in\chi\}$.
\end{itemize}

The most important result for such construction is the truth
lemma.
\begin{mylma}[Truth lemma]\label{lma2}
For any wff $\varphi$ and DBF$_n^c$-MCS $\chi$, we have
$w_\chi\models_{M^*}\varphi$ iff $\varphi\in\chi$.
\end{mylma}
{\bf Proof}: By induction on the structure of the wff, the only
interesting case is the wff of the form $[I]\psi$ for some
$I\in{\cal I}$. By definition, $w_\chi\models_{M^*}[I]\psi$ iff
for all $w_{\chi'}\in {\cal R}^*_I(w_\chi)$,
$w_{\chi'}\models_{M^*}\psi$ iff for all $\chi/I\subseteq\chi'$,
$\psi\in\chi'$ (by induction hypothesis) iff
$\chi/I\cup\{\neg\psi\}$ is inconsistent iff $[I]\psi\in\chi$ when
$[I]$ is a normal modal operator\cite{che}. However, by the axioms
P and G1, rules MP and Gen, and propositions \ref{prop1}.2 and
\ref{prop1}.4, both kinds of modal operators $[O]$ and $[G]$ are
normal ones. $\Box$

Since every DBF$_n^c$-MCS contains all wffs provable in DBF$_n^c$,
by the truth lemma, all provable wffs are valid in $M^*$.
Furthermore, by axiom G2, each ${\cal R}^*_{\{i\}}$ is serial for
$1\leq i\leq n$. Thus $M^*$ is indeed a pseudo model. If $\varphi$
is DBF$_n^c$-consistent, then there exists an MCS $\chi$
containing $\varphi$, so by the truth lemma,
$w_\chi\models_{M^*}\varphi$, i.e., $\varphi$ is pseudo
DBF$_n^c$-satisfiable. This proves the first part of lemma
\ref{lma1}.

Note that if $|{\cal I}|=m$, then a pseudo model is in fact a
model for the multi-agent epistemic logic K$_m$\cite{fag}. The
logic K$_m$ has $m$ modal operators corresponding to the knowledge
or belief of $m$ independent agents. Admittedly, $m$ may be a very
large number, however, it does not matter for the current purpose.
What is important is that it can be shown that without loss of
generality, we can assume a pseudo model is tree-like. The detail
definition of a tree-like model and the proof that each pseudo
model can be ``unwound'' into a tree-like one verifying the same
set of valid wffs are rather technical and can be found in
(\cite{fag1},pp.354) and (\cite{fag},Exercise 3.30). What is
needed here is the property that in a tree-like model, if
$I\not=J$, then ${\cal R}^*_I\cap{\cal R}^*_J=\emptyset$.

Thus, from now on, we can assume that if $\varphi$ is
DBF$_n^c$-consistent, then $\varphi$ is pseudo
DBF$_n^c$-satisfiable in a tree-like model $M^*=(W,({\cal
R}^*_I)_{I\in{\cal I}}, V)$. The next step is to construct a
DBF$_n^c$ model $M=(W,({\cal R}_i)_{1\leq i\leq n}, V)$ from $M^*$
by defining ${\cal R}_i=\bigcup_{i\in G}{\cal R}^*_G$. Note that
${\cal R}_i$ is serial since ${\cal R}_i\supseteq{\cal
R}^*_{\{i\}}$ which is serial by the definition of pseudo models.
From the definition, we can prove the following lemma.
\begin{mylma}\label{lma3}
For any $w\in W$ and wff $\varphi$, $w\models_{M^*}\varphi$ iff
$w\models_{M}\varphi$.
\end{mylma}
{\bf Proof}: By induction on the structure of $\varphi$, the basis
and classical cases are easy since both models have the same truth
assignment function $V$. For the modal cases, if
$\varphi=[G]\psi$, then $w\models_{M}[G]\psi$ iff for all
$u\in\bigcap_{i\in G}{\cal R}_i(w)$, $u\models_{M}\psi$ iff for
all $u\in\bigcup_{G'\supseteq G}{\cal R}^*_{G'}(w)$,
$u\models_{M^*}\psi$ (by the definition of ${\cal R}_i$, the
tree-likeness of $M^*$, and the induction hypothesis) iff
$w\models_{M^*}\bigwedge_{G'\supseteq G}[G']\psi$ (by definition
of satisfaction in pseudo model $M^*$) iff $w\models_{M^*}[G]\psi$
(since axiom G3 is valid in $M^*$).

If $\varphi=[O]\psi$, then since proposition \ref{prop1}.1 is both
valid in $M^*$ (by definition of pseudo models) and $M$ (by
soundness), and by the case for modal operators $[G]$, we can find
a $j$ such that $w$ satisfies the wff
$\neg[G_j]\bot\wedge[G_{j+1}]\bot$ (or just $\neg[G_j]\bot$ in
case of $j=|\delta(O)|$) in both $M$ and $M^*$, so it can be shown
that $w\models_{M}[O]\psi$ iff $w\models_{M}[G_j]\psi$ iff
$w\models_{M^*}[G_j]\psi$ iff $w\models_{M^*}[O]\psi$. $\Box$

This finishes the proof for the second part of lemma \ref{lma1}
and by combining the two parts, we have proved the completeness
theorem for DBF$_n^c$.

\subsection{Proof of Theorem \ref{thm2}}
The proof is very analogous to the previous one. What is different
is that we do not have a counterpart for proposition \ref{prop1}.1
in the system DBF$_n^s$. First, a pseudo DBF$_n^s$ structure is
analogously defined as a tuple $(W,({\cal
R}_\Omega)_{\emptyset\not=\Omega\subseteq{\cal TO}_n}, V)$ and it
is required that ${\cal R}_{\{i\}}$ is serial for all $1\leq i\leq
n$. Then a pseudo DBF$_n^s$ model is a pseudo DBF$_n^s$ structure
in which all wffs provable in DBF$_n^s$ are valid.

We still have to prove the following lemma.
\begin{mylma}\label{lma4}
\begin{enumerate}
\item If $\varphi$ is DBF$_n^s$-consistent, then $\varphi$ is pseudo
DBF$_n^s$-satisfiable.
\item If $\varphi$ is pseudo DBF$_n^s$-satisfiable, then it is DBF$_n^s$-satisfiable.
\end{enumerate}
\end{mylma}

The first part of the lemma is proved exactly in the same way as
in lemma \ref{lma1}. It can also be obtained  that if $\varphi$ is
DBF$_n^s$-consistent, then $\varphi$ is pseudo
DBF$_n^s$-satisfiable in a tree-like model $M^*=(W,({\cal
R}^*_\Omega)_{\emptyset\not=\Omega\subseteq{\cal TO}_n}, V)$.

However, the proof of the second part is somewhat different. Let
us define the {\em level \/} of a modal operator $\Omega$ as
$l(\Omega)=\max_{O\in\Omega}|\delta(O)|$ and the {\em length \/}
of $\Omega$ as $\sharp(\Omega)=$ the number of elements $O$ in
$\Omega$ such that $|\delta(O)|=l(\Omega)$. Then we define a
function $Ag^*:W\times (2^{{\cal
TO}_n}-\{\emptyset\})\ra(2^{\{1,2,\ldots,n\}}-{\emptyset})$ from
the model $M^*$ by
\[Ag^*(w,\Omega)=\left\{\begin{array}{ll}
 \bigcup_{O\in\Omega}Ag^*(w,\{O\}) & {\rm if} \;|\Omega|>1,\\
Ag^*(w,\{O\})\cup\{i\}& {\rm if} \; \Omega=\{O>i\} \; {\rm and} \;
w\models_{M^*}\neg[\{O,i\}]\bot,\\ Ag^*(w,\{O\})& {\rm if} \;
\Omega=\{O>i\} \; {\rm and} \; w\models_{M^*}[\{O,i\}]\bot,\\
\{i\}& {\rm if} \; \Omega=\{i\},
\end{array}
\right.\] Note that since $Ag^*(w,\Omega)$ is a subset of agents,
it can also be used as a modal operator of level 1. We can now
construct a DBF$_n^s$ model $M=(W,({\cal R}_i)_{1\leq i\leq n},
V)$ from $M^*$ such that ${\cal R}_i=\bigcup_{l(\Omega)=1,
i\in\Omega}{\cal R}^*_\Omega$.
\begin{mylma}\label{lma5}
\begin{enumerate}
\item For all $w\in W$, modal operators $\Omega$, and wffs $\varphi$, we have
$w\models_{M^*}[\Omega]\varphi\equiv[Ag^*(w,\Omega)]\varphi$.
\item ${\cal R}_\Omega(w)=\bigcup\{{\cal R}^*_{\Omega'}(w)\mid l(\Omega')=1\wedge
Ag^*(w,\Omega)\subseteq\Omega'\}$ for all $w\in W$ and modal
operators $\Omega$, where ${\cal R}_\Omega$ is defined in section
\ref{sec4}.
\end{enumerate}
\end{mylma}
{\bf Proof}:
\begin{enumerate}
\item By induction on the level of $\Omega$:
\begin{description}
\item The basis case $l(\Omega)=1$: then by definition,
$Ag^*(w,\Omega)=\Omega$, so the result holds trivially.
\item Assume the result holds for all $\Omega$ such that
$l(\Omega)\leq k$,
\item $l(\Omega)=k+1$: by induction on the length of $\Omega$:
\begin{description}
\item $\sharp(\Omega)=1$: let $\Omega=\{O>i\}\cup\Omega_1$, where
$l(\Omega_1)\leq k$, then $Ag^*(w,\Omega)=Ag^*(w,\{O>i\})\cup
Ag^*(w,\Omega_1)=Ag^*(w,\{O\})\cup
Ag^*(w,\Omega_1)\cup\{i\}=Ag^*(w,\Omega_2)$ if
$w\models_{M^*}\neg[\{O,i\}]\bot$ and $=Ag^*(w,\{O\})\cup
Ag^*(w,\Omega_1)=Ag^*(w,\Omega_3)$ if
$w\models_{M^*}[\{O,i\}]\bot$, where
$\Omega_2=\{O,i\}\cup\Omega_1$ and $\Omega_3=\{O\}\cup\Omega_1$.
Since $l(\Omega_2)=l(\Omega_3)=k$, then by induction hypothesis,
we have
$w\models_{M^*}[\Omega_i]\varphi\equiv[Ag^*(w,\Omega_i)]\varphi$
for $i=2,3$, so by axioms O1' and O2' (recall that all axioms are
valid in a pseudo model),
$w\models_{M^*}[\Omega]\varphi\equiv[Ag^*(w,\Omega)]\varphi$ no
matter whether $w\models_{M^*}[\{O,i\}]\bot$ or not.
\item Assume the result holds for all $\Omega$ such that
$\sharp(\Omega)\leq t$:
\item $\sharp(\Omega)=t+1$: the induction step is completely the
same as in the basis case except that
$l(\Omega_2)=l(\Omega_3)=k+1$ but
$\sharp(\Omega_2)=\sharp(\Omega_3)=t$.
\end{description}
\end{description}
\item By induction on $l(\Omega)$:
\begin{description}
\item $l(\Omega)=1$: then $Ag^*(w,\Omega)=\Omega$ and by
definition in section \ref{sec4}
\begin{eqnarray*}
{\cal R}_\Omega(w)& = &\bigcap_{i\in\Omega}{\cal R}_i(w)\\
 &=&\bigcap_{i\in\Omega}\bigcup\{{\cal R}^*_{\Omega'}(w)\mid l(\Omega')=1\wedge
i\in\Omega'\}\\
 &=&\bigcup\{{\cal R}^*_{\Omega'}(w)\mid l(\Omega')=1\wedge
\Omega\subseteq\Omega'\}\\
 &=&\bigcup\{{\cal R}^*_{\Omega'}(w)\mid l(\Omega')=1\wedge
Ag^*(w,\Omega)\subseteq\Omega'\}
\end{eqnarray*}
\item Assume the result holds for $l(\Omega)\leq k$.
\item $l(\Omega)=k+1$: there are two cases
\begin{description}
\item $|\Omega|=1$: let $\Omega=\{O>i\}$, then by definition in
section \ref{sec4}, we have
\[{\cal R}_\Omega(w)={\cal R}_{O>i}(w)=\left\{\begin{array}{ll}
 {\cal R}_O(w) & {\rm if} \;{\cal R}_O(w)\cap{\cal R}_i(w)=\emptyset,\\
{\cal R}_O(w)\cap {\cal R}_i(w)& {\rm otherwise},
\end{array}
\right.\] where by the induction hypothesis and the definitions of
$Ag^*$ and ${\cal R}_i(w)$,
\[{\cal R}_O(w)=\bigcup\{{\cal R}^*_{\Omega'}(w)\mid l(\Omega')=1\wedge
Ag^*(w,\{O\})\subseteq\Omega'\}\]
\[{\cal R}_O(w)\cap{\cal R}_i(w)=\bigcup\{{\cal R}^*_{\Omega'}(w)\mid l(\Omega')=1\wedge
Ag^*(w,\{O,i\})\subseteq\Omega'\}.\] On the other hand, by the
result of first part, let $\Omega_1=Ag^*(w,\{O,i\})$, then
$w\models_{M^*}[\{O,i\}]\bot$ iff $w\models_{M^*}[\Omega_1]\bot$
iff $w\models_{M^*}[\Omega']\bot$ for all $\Omega'$ such that
$l(\Omega')=1$ and $\Omega_1\subseteq\Omega'$ (by axiom V3) iff
${\cal R}^*_{\Omega'}(w)=\emptyset$ for all such $\Omega'$ iff
${\cal R}_O(w)\cap{\cal R}_i(w)=\emptyset$. Thus, by the
definition of $Ag^*$,
\[Ag^*(w,\Omega)=\left\{\begin{array}{ll}
Ag^*(w,\{O\})& {\rm if} {\cal R}_O(w)\cap{\cal R}_i(w)=\emptyset\\
Ag^*(w,\{O,i\})& {\rm otherwise}
\end{array}
\right.\] and the result follows immediately.
\item  $|\Omega|>1$: by definition
\begin{eqnarray*}
{\cal R}_\Omega(w) &= &\bigcap_{O\in\Omega}{\cal R}_O(w)\\
 &=&\bigcap_{O\in\Omega} \bigcup\{{\cal R}^*_{\Omega'}(w)\mid l(\Omega')=1\wedge
Ag^*(w,\{O\})\subseteq\Omega'\}\\
 &=&\bigcup\{{\cal R}^*_{\Omega'}(w)\mid l(\Omega')=1\wedge \bigcup_{O\in\Omega}Ag^*(w,\{O\})\subseteq\Omega'\}\\
 &=&\bigcup\{{\cal R}^*_{\Omega'}(w)\mid l(\Omega')=1\wedge
Ag^*(w,\Omega)\subseteq\Omega'\}
\end{eqnarray*}
\end{description}
\end{description}
\end{enumerate}
$\Box$

Finally, we can prove the counterpart of lemma \ref{lma3} for
DBF$_n^s$
\begin{mylma}\label{lma6}
For any $w\in W$ and wff $\varphi$, $w\models_{M^*}\varphi$ iff
$w\models_{M}\varphi$.
\end{mylma}
{\bf Proof}: By induction on the structure of $\varphi$, the only
interesting case is $\varphi=[\Omega]\psi$,
\begin{eqnarray*}
w\models_{M^*}[\Omega]\psi & \Leftrightarrow &
w\models_{M^*}[Ag^*(w,\Omega)]\psi\mbox{ \rm (lemma
 \ref{lma5}.1)}\\
  & \Leftrightarrow & w\models_{M^*}[\Omega']\psi \mbox{ \rm for
  all } \Omega' \mbox{ \rm such that } l(\Omega')=1 \mbox{ \rm and }
Ag^*(w,\Omega)\subseteq\Omega'\mbox{ \rm (V3)}\\
 & \Leftrightarrow & u\models_{M^*}\psi, \forall u\in \bigcup\{{\cal R}^*_{\Omega'}(w)\mid l(\Omega')=1\wedge
Ag^*(w,\Omega)\subseteq\Omega'\}\\
 & \Leftrightarrow & u\models_{M}\psi, \forall u\in{\cal
 R}_{\Omega}(w)\mbox{ \rm (induction hypothesis and lemma
 \ref{lma5}.2)}\\
 & \Leftrightarrow & w\models_{M}[\Omega]\psi
\end{eqnarray*}$\Box$

This completes the proof of the second part of lemma \ref{lma4}
and the completeness theorem for DBF$_n^s$.

\end{document}